# A persistent accuracy ceiling in automated verbal deception detection

Riccardo Loconte [1]*, Jonas Festor [1], Zane Fatjanova [1], Mariam Bolkvadze [1], Bennett Kleinberg[1,2]

[1] Department of Methodology and Statistics, Tilburg University; Tilburg, Netherlands.
[2] Department of Security and Crime Science, University College London; London, UK.
* Corresponding author. E-mail: r.loconte@tilburguniversity.edu

## Abstract

Automated methods have been proposed to overcome the limitations of human verbal deception detection, but evidence remains fragmented across disciplines. We systematically reviewed 25 years of research (289 reports, 6,136 classification models) and meta-analyzed 3,653 models nested within 97 datasets. Pooled accuracy was 74.4% (95% CI: 71.2%-77.4%) with substantial heterogeneity. Accuracy was driven by methodological quality (ground truth, data source, class balance, evaluation procedure) more than by model complexity: the adoption of embeddings and large language models has not translated into improved predictive performance. Only 12.46% of reports used data with verifiable ground-truth, and only 23.96% of models were evaluated on independent data. The pooled accuracy aligns with meta-analyses of manual approaches, suggesting a ceiling of 70-75%, unlikely to be lifted by current research conventions.

## Main Text

Deception is the "*psychological process by which one individual deliberately attempts to convince another person to accept as true what the liar knows to be false, typically […] to gain some type of benefit or to avoid loss*" (*1*). Deceptive intent and communication occur daily and across a wide range of contexts (*2–5*). While typical everyday white lies are trivial, detecting deception carries significant implications in contexts such as legal and forensic settings, where the aim is to evaluate a suspect's credibility during investigative interviews (*6*). False or distorted information misleads investigations, delays case resolution, and may result in wrongful convictions, undermining justice and public trust in the legal system (*7*, *8*). In financial services, deception translates into fraud when individuals or organizations manipulate transaction reports to gain unfair and unlawful economic advantage (*9*). In online platforms, deceptive practices erode trust by posting fake reviews that distort consumer perceptions and unfairly promote certain businesses (*10*, *11*).

Despite its social relevance, meta-analytical research on deception detection abilities has shown that humans, including professionals, perform poorly, often not better than the chance level (*12–14*). In response, research has focused on improving deception accuracy through strategies that elicit verbal cues from strategic interviewing (*15*, *16*) or structured content-based analysis (*17–20*), through cues such as the number of details or inconsistencies within a statement. With the development of dictionary-based approaches to automatically extracting variables from textual data (*21*), computer-aided detection of verbal deception has received more attention (*22–24*), and early meta-analytic evidence has highlighted its potential (*25*). However, over the last two decades, the analysis of textual data has advanced with major developments in machine learning (ML), such as the introduction of deep learning architectures composed of multiple processing layers that learn how to represent data at various levels of abstraction (*26*), and natural language processing (NLP), including embeddings for text representation (*27–30*) and language models for text generation (*30–32*), enabling novel approaches to handle textual data. Consequently, these advances have opened new avenues for interdisciplinary research on computational models for detecting deception in text – a field known as *automated verbal deception detection* (Fig. 1).

In this analytical review, we show the current state of automated verbal deception detection *i)* by mapping the research practices adopted in the field and *ii*) by meta-analyzing the performance of deception classifiers and important moderators. We systematically review and analyze the work published in the past 25 years of research, presenting a near-complete view of the field that includes 289 reports and 6,136 classification models. This study is the largest synthesis of automated verbal deception research to date, extending and updating previous systematic reviews (*33*, *34*) and meta-analysis (*25*) in the field. Details about the literature screening, data coding, and the meta-analytic plan are reported in Supplementary Information. Table 1 presents a glossary of the most important terms used in this work.

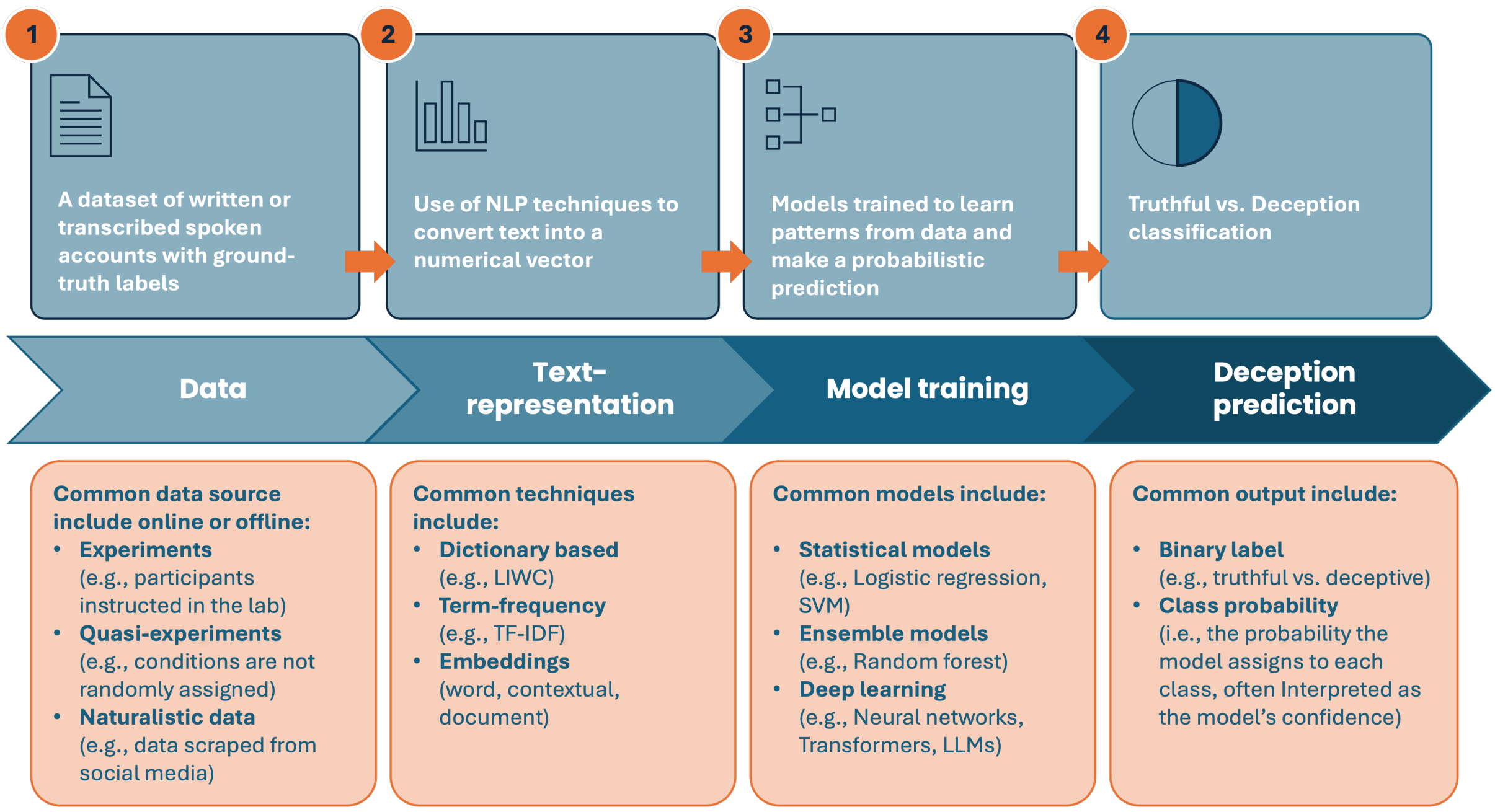

**Fig 1. Pipeline for automated verbal deception detection.**
A typical pipeline involves (1) collecting a dataset of written or transcribed statements with ground-truth labels through (quasi-)experiments or naturalistic research designs, (2) using NLP techniques to convert the text into numerical vectors, (3) training machine learning models on these vectors, and (4) using the trained models to classify a statement as truthful or deceptive.

**Table 1.** Glossary of overarching concepts.

| LEVEL | | DEFINITION |
|---|---|---|
| Origin | **Online sources** | Data collected from digital environments (e.g., web platforms, social media, online surveys). |
| | **Offline sources** | Data collected in vivo, either in controlled or natural settings (e.g., lab experiments, field studies, face-to-face interviews). |
| Research design | **Experiments** | Studies which actively manipulate variables under controlled conditions to establish cause-and-effect relationships. |
| | **Quasi-experiment** | Studies that examine causal relationships without full randomization or control |
| | **Naturalistic** | Information collected from real-life settings without experimental manipulation (i.e., "found data"; (*35*)) |
| Ground truth | **Clear and verifiable** | The ground truth is fully transparent and verifiable (e.g., object descriptions, videos, mock crimes). |
| | **Clear but not verifiable** | The ground truth is clearly defined through experimental manipulation, but individual statements cannot be verified. Ground truth is based on participants' compliance. |
| | **Directly inferred** | The ground truth is inferred from plausible proxy variables indirect indicators (e.g., deception is inferred from judges' decisions in trial-hearing transcripts). |
| | **Indirectly inferred** | The ground truth is inferred from unreliable proxy variables, based on weak inference methods, reducing confidence in its reliability (e.g., deception is inferred from human impressions). |
| Text-representation | **Linguistic and statistical features** | Text is represented as a numerical vector of counts or (or proportion) of features derived from dictionary-based methods or surface-level textual properties (e.g., LIWC, part-of-speech tags, named-entity recognition). |
| | **Term frequencies** | Text-representation as a vector of *n*-gram frequencies ignoring word order (e.g., bag-of-words, term frequency-inverted document frequency). |
| | **Embeddings** | Dense vector representations of text that capture the semantic of textual data (e.g., word2vec, BERT, Doc2vec, ModernBERT) |
| | **Topic and latent semantic features** | Text-representation based on latent themes (e.g., topic modelling, latent Dirichlet allocation). |
| | **Hybrid approaches** | Combinations of text-representation classes (e.g., LIWC + TF-IDF). |
| Model categories | **Statistical learning models** | Models that require explicit text representations and rely on statistical principles (e.g., logistic regression, naïve bayes, and support vector machine). |
| | **Ensemble models** | Models that combine predictions from multiple models (e.g., bagging, boosting, voting, Random Forest, Xgboost, lightGBM). |
| | **Neural networks** | Models that learn representations through interconnected layers and artificial neurons (e.g., Feedforward networks, Convolutional neural networks, Recurrent neural networks, Long-term short memory) |
| | **Transformer-based language models** | Deep learning models pretrained on large corpora capturing long-range, contextual dependencies via self-attention mechanisms (e.g., Bidirectional Encoder Representation from Transformers, BERT; Robustly Optimized BERT Pretraining Approach, RoBERTa) |
| | **Large language models** | Large-scale, transformer-based models using autoregressive next-token prediction capable of performing few-shot or zero-shot classification through prompt engineering (e.g., GPT-4.5, FLAN-T5, Llama-2) |
| Evaluation procedure | **In-sample evaluation** | Model performance is tested on the same data used for training |
| | **Hold-out test set** | The model is tested on a random subset of the dataset that was not used during training. |
| | **Within cross-validation** | The model is iteratively trained on $k-1$ folds and tested on the remaining one, with the overall performance computed as the average of the evaluation metrics across all k iterations. |
| | **Tested on different data** | The model is trained on one dataset and tested on a separate, independent dataset. |

## The performance ceiling around 70-75%

The meta-analysis revealed that the pooled estimated accuracy in detecting deceptive and truthful statements of machine learning models predicting verbal deception is 74.4% (95% CI: 71.2%-77.4%) (Fig. 2). Importantly, this estimate is adjusted for class imbalance and refers to balanced classes (chance level = 50%). In imbalanced datasets, accuracy rose only modestly (i.e., 77% accuracy for a majority-class proportion of 0.60) and it even fell below the accuracy of a trivial classifier that always predicts the majority class when class distributions were highly skewed (i.e., majority class equals 0.90, the average prediction is 83.6%). These findings suggest that models appear to underexploit class imbalance, increasingly so as imbalance grows.

Notably, this pooled estimate exceeds the performance of earlier dictionary-based methods (e.g., the Linguistic Inquiry and Word Count software; LIWC (*21*, *36*, *37*)) used to examine the effectiveness of single verbal cues, which achieved 55.2% accuracy[1] (Hedge's $g = 0.26$) (*25*). Similarly, the verifiability approach - which also relies on a single cue, namely the presence of verifiable details - yields a comparable estimate ($g = 0.49$, corresponding to an accuracy of 59.7%) (*38*, *39*). However, in most automated verbal deception detection research, deception is detected from a combination of multiple cues. When multiple cues are effectively combined, deception can theoretically be detected with up to 67.86% accuracy, as suggested in other meta-analytical research (*40*). Therefore, we can speculate that the performance improvement we observed in our pooled estimate is attributable not only to more advanced techniques for text representation but also to a more effective outcome prediction relying on a combination of multiple verbal cues. Yet, this pooled performance converges closely with other meta-analytic estimates from theory-driven, manually coded approaches, such as content-based analyses like the Criteria-Based Content Analysis and Reality Monitoring (which yield 70% accuracy) (*17, 41*) and the cognitive load elicitation approach (yielding to 71% accuracy against 56% for standard interviewing) (*16*). This convergence at a 70-75% accuracy across conceptually unrelated approaches suggests an accuracy ceiling in verbal deception detection.

But why does this ceiling persist in light of major advancements in the fields contributing to artificial intelligence, such as NLP and machine learning? Our moderator analysis (Table 2; see also the Supplementary Information) suggests that two sets of factors, namely methodological quality (e.g., ground truth and data source) and computational choices (e.g., text representation, model architecture, and evaluation procedures), were significantly associated with performance, with the first showing larger effects than the second. The next sections discuss their impact in more detail.

---

[1] Conversion of Hedges' $g$ to classification accuracy was done with the formula accuracy = $\Phi(g/2)$, where $\Phi$ is the standard normal cumulative distribution function, assuming that the number of truthful and deceptive statements is balanced.

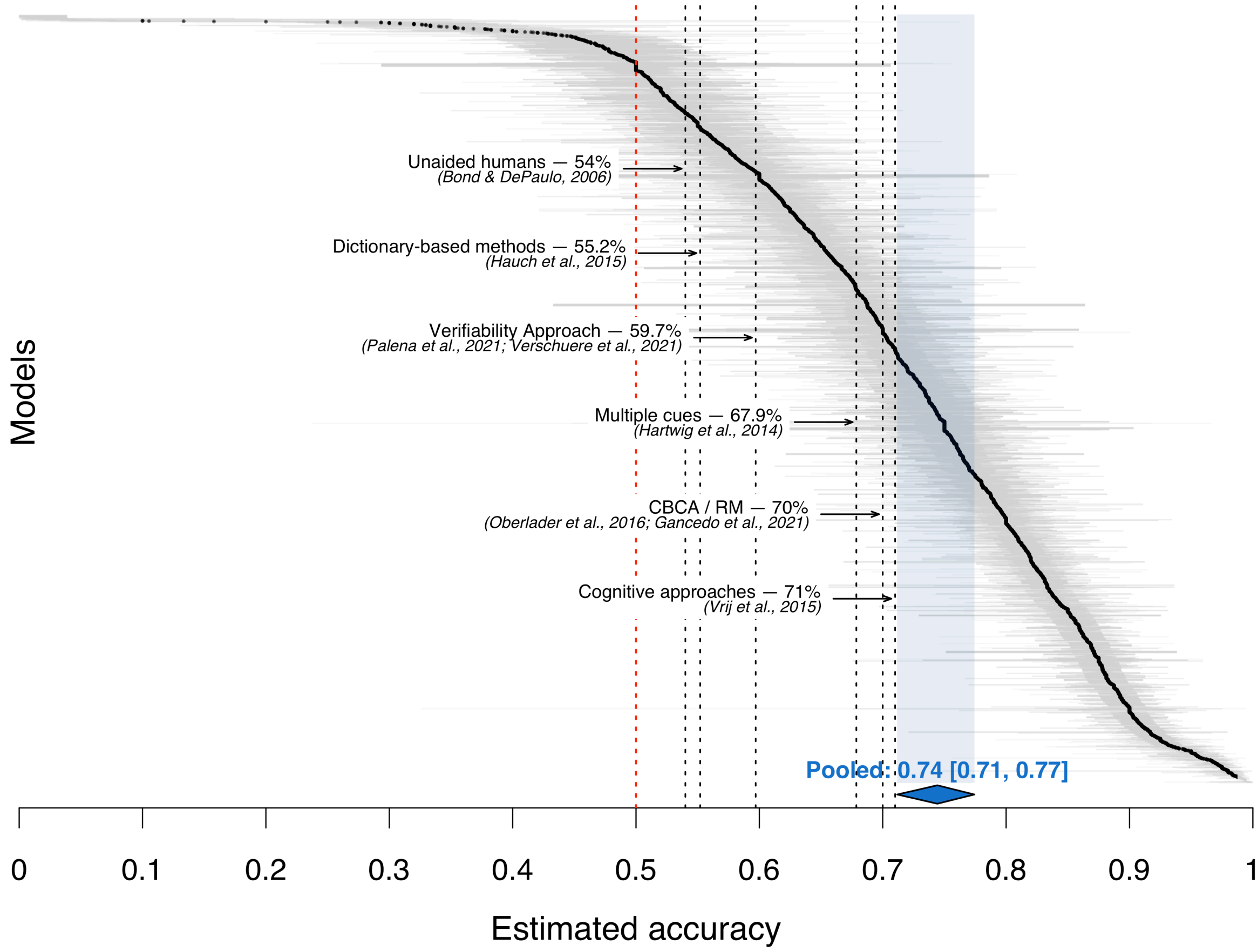

**Fig. 2. Caterpillar plot of model-level accuracy estimates.**
Each point represents a model's estimated accuracy, ordered from lowest to highest and back-transformed from the logit scale ($n$ = 3,653). The red dashed vertical line indicates the chance level when truthful and deceptive classes are balanced. The light blue vertical bar represents, together with the blue diamond, the 95% CI of the pooled estimated accuracy.

## The methodological barriers of the ceiling effect

### *Ground truth and the internal validity problem*

Our findings indicate that methodological rigor in operationalizing ground-truth is generally low in the field. The term ground truth originates from remote sensing science and geography and refers to the process of verifying digitized information (e.g., images) about a landmark captured from the air (*42*). By extension, in deception detection research, ground truth has been defined as what we know to be true, as confirmed by empirical evidence (e.g., direct observation and measurement), rather than by inference or conjecture (*43*). Establishing a reliable ground truth is fundamental to internal validity because it ensures that observed effects can be attributed to actual deception rather than measurement error or ambiguous labeling (*44*). After coding the quality of ground truth in six levels (Table 1), findings revealed that only 12.46% ($n$=36) of reports employed a *clear and verifiable ground truth* by using tasks that allow for full verification of statements, such as object description, video recollection, or mock crime tasks. In contrast, the largest share of reports (38.06%, $n$=110) used a *clear but non-verifiable ground truth*, typically via experimental manipulations that assign participants to truthful vs. deceptive conditions, with little control over what participants actually fabricate. This occurs despite evidence that people may unwittingly embed deceptive details into their truthful accounts (*45*) or draw on similar past truthful experiences to fabricate (*46*), undermining the assumption that manipulating instructions is sufficient to determine a valid ground truth. Even more concerningly, a substantial number of reports (25.95%, $n$=75) worked with a level of ground truth that was inferred by employing plausible or unreliable *ad hoc* criteria (see *directly* and *indirectly inferred ground-truth* definitions in Table 1). Examples include manual annotation of deception by crowd raters, expert judges, or research assistants, largely ignoring evidence that such human-inferred annotation is typically error-prone (*12–14*).

Our meta-analysis further shows that ground truth operationalization is systematically associated with reported performance (Table 2): relative to a clear and verifiable ground truth, a non-verifiable ground truth may inflate accuracy ($\beta = 0.48$, 95% CI [0.33, 0.63], $p < .001$) by +5.7 percentage points, possibly because experimental assignment more cleanly separates truth from deception. In contrast, both directly inferred ($\beta = -0.35$, [−0.54, −0.15], $p = .001$, -5.2 percentage points) and indirectly inferred ground truth ($\beta = -3.71$, [−3.92, −3.49], $p < .001$, -69.4 percentage points) blur the separation of truth and deception, consistent with increased label noise. Ground truth rigor is the field's elephant in the room: poor ground truth weakens internal validity and, as the moderator estimates above show, systematically shifts reported performance in both directions. Strengthening ground truth must be a fundamental requirement for future research.

Finally, research on verbal deception faces a persistent trade-off between ecological and internal validity. Naturalistic designs better capture deception as it occurs in real-world contexts where it is self-initiated, context-dependent, and based on realistic truth-lie base rates. Examples include data collected on online platforms, such as fake reviews (*47*, *48*) or fake job posts (*49–51*), or in real-life settings, such as transcripts of trial hearings (*52*). Yet, naturalistic designs often rely on inferred rather than verified ground-truth labels (*44*, *53*), compromising their internal validity. Controlled experiments, by contrast, permit stronger verification of truthfulness and greater experimental control, but risk oversimplifying deception and hence limit generalizability (*44*, *54*). Thus, an inherent, unresolved tension remains between studying deception realistically and confidently establishing whether and where deception occurred.

**Table 2.** Beta coefficients, standard error, 95% confidence interval, *z*-value, and p-value of all moderators.

| Predictors | *n* | #ES | Estimates† | *SE* | 95% CI | *z* | *p* | Pred. Acc. (95% CI) ‡ |
|---|---|---|---|---|---|---|---|---|
| (Intercept) | | | 0.22 | 0.40 | -0.56, 1.00 | 0.56 | 0.5759 | |
| ***Majority class*** (Ref.: 0.50) | 161 | 3653 | 1.30 | 0.02 | 1.26, 1.35 | 59.95 | < .001*** | 79.4 (74.5, 83.2) |
| ***Source*** | | | | | | | | |
| (Ref.: Offline experiment) | 21 | 323 | - | - | - | - | - | 62.0 (45.4, 75.6) |
| Offline quasi-experiment | 1 | 2 | 0.92 | 0.46 | 0.02, 1.83 | 2.00 | 0.046* | 77.4 (70.0, 83.1) |
| Offline naturalistic data | 42 | 365 | 0.78 | 0.59 | -0.37, 1.93 | 1.33 | 0.183 | 75.4 (60.4, 85.2) |
| Online experiment | 26 | 1168 | 1.17 | 0.46 | 0.27, 2.07 | 2.55 | 0.011* | 80.5 (74.3, 85.4) |
| Online quasi-experiment | 57 | 1464 | 1.59 | 0.46 | 0.68, 2.49 | 3.45 | < .001*** | 84.7 (79.8, 88.6) |
| Online naturalistic data | 33 | 300 | 1.42 | 0.46 | 0.51, 2.32 | 3.07 | 0.002 ** | 83.1 (77.7, 87.4) |
| Mixed | 6 | 31 | 3.88 | 0.78 | 2.35, 5.40 | 4.99 | < .001*** | 96.5 (91.1, 98.8) |
| ***Ground truth*** | | | | | | | | |
| (Ref.: Clear and verifiable) | 21 | 294 | - | - | - | - | - | 82.2 (76.3, 86.7) |
| Clear but not verifiable | 93 | 2673 | 0.48 | 0.08 | 0.33, 0.63 | 6.33 | < .001*** | 87.9 (83.5, 90.9) |
| Directly inferred | 39 | 389 | - 0.35 | 0.10 | -0.54, -0.15 | -3.47 | 0.001*** | 77.0 (69.8, 82.8) |
| Indirectly inferred | 32 | 297 | - 3.71 | 0.11 | -3.92, -3.49 | -34.36 | < .001*** | 12.8 (9.0, 18.3)§ |
| ***Text-representation*** | | | | | | | | |
| (Ref.: Linguistic and statistical features) | 72 | 620 | - | - | - | - | - | 78.0 (72.9, 82.2) |
| Term frequency | 77 | 1048 | 0.25 | 0.005 | 0.24, 0.26 | 54.62 | < .001*** | 81.1 (76.4, 84.6) |
| Topic and latent semantic features | 2 | 20 | 0.11 | 0.01 | 0.08, 0.13 | 8.75 | < .001*** | 79.4 (74.4, 83.3) |
| Embeddings | 70 | 1161 | 0.12 | 0.006 | 0.11, 0.13 | 19.68 | < .001*** | 79.5 (74.6, 83.4) |
| Hybrid | 52 | 804 | 0.33 | 0.005 | 0.32, 0.34 | 67.46 | < .001*** | 81.9 (77.4, 85.4) |
| ***Model category*** | | | | | | | | |
| (Ref.: Statistical learning models) | 116 | 1789 | - | - | - | - | - | 79.9 (74.7, 83.6) |
| Ensemble models | 54 | 484 | 0.04 | 0.005 | 0.04, 0.05 | 8.96 | < .001*** | 80.4 (75.3, 84.1) |
| Neural networks | 70 | 576 | 0.03 | 0.005 | 0.02, 0.04 | 5.67 | < .001*** | 80.2 (75.1, 83.9) |
| Transformer-based models | 31 | 293 | 0.32 | 0.01 | 0.30, 0.34 | 31.41 | < .001*** | 83.4 (79.0, 86.6) |
| Large language models | 12 | 511 | -0.09 | 0.02 | -0.13, -0.06 | -5.49 | < .001*** | 78.7 (73.4, 82.6) |
| **Evaluation** | | | | | | | | |
| (Ref.: Within-cross validation) | 100 | 1758 | - | - | - | - | - | 81.1 (76.4, 84.7) |
| Hold out test-set | 61 | 1074 | 0.06 | 0.01 | 0.03, 0.09 | 4.25 | < .001*** | 81.7 (77.1, 85.2) |
| Different dataset | 30 | 821 | -0.38 | 0.01 | -0.40, -0.36 | -39.04 | < .001*** | 76.4 (70.7, 80.7) |

* $p < 0.05$; ** $p < 0.01$; *** $p < 0.001$.
Abbreviations: *n* = number of reports; #ES= number of effect sizes; CI = confidence intervals
† Estimates represent β coefficients contrasted against the reference level (logit scale).
‡ Predicted accuracy represents average marginal predictions on the proportion scale, averaged over the observed distribution of all other moderators. Confidence intervals were estimated by parametric simulation (*n*=5000) from the fitted model using the estimated variance-covariance matrix of the regression coefficients.
§ This estimate should be interpreted with caution because the indirectly inferred ground-truth is concentrated almost entirely within the online naturalistic data source and almost not present in offline research designs. Hence, the model cannot fully separate the effects of the ground-truth and the source, so this marginal prediction is confounded by dataset-level differences.

### ***Data source matters***

Data sources varied substantially across reports, ranging from online and offline data collected through experiments, quasi-experiments, and naturalistic research designs (see Table S1 and S3). Typically, offline sources yielded smaller datasets (e.g., from small-scale controlled experiments) than online sources (e.g., large-scale scraped data) (Table S3-S4), with a real impact on model performance. For example, our meta-analysis showed that analyses based on data collected in online experiments resulted in a higher accuracy than those conducted on data from offline experiments by 18.5 percentage points, potentially because online data collection can readily provide larger training sets, which in turn yield more accurate models. But while this heterogeneity is likely explained by more readily accessible large-scale online data, it may also reflect inter-disciplinary differences (*34*). Psychology and computer science often differ in research goals, methodology, and data accessibility, with psychology having traditionally favored controlled lab experiments to study causal relationships between variables, and computer science having largely aimed to rely on available large-scale datasets with a focus on predictive modeling (*34*, *57–59*). Overall, source-related differences in data scale and accessibility appear to shape predictive performance, but the ability to collect high-quality data at scale is itself shaped by disciplinary conventions, highlighting the value of greater methodological exchange between psychology and computer science.

### *The narrow horizon of deception*

Deception was mostly examined in terms of fabrication (*n*=184, 63.67% of reports), in English datasets (*n*=237, 82.01%)[2], and via tasks focused on the production of fake reviews (*n*=120, 41.52%) or past experiences (*n*=46, 15.92%)[3] (see Table S2). Focusing on fabrication may facilitate research in magnifying differences between truthful and deceptive statements, but it ignores that real-life deception has been shown to occur mostly in more nuanced forms, such as embedded lies (*5*, *45*, *46*). Further, the English-centric focus may be a consequence of English-language data being more readily available and might further reinforce this dominance. To date, deception research has remained relatively narrow in scope, limiting its external validity and generalizability to real-world settings where deception is typically subtle (*45*, *46*), context-dependent (*55*), and culturally mediated (*56*). Future research needs to engage more in experimental and naturalistic designs that capture the full spectrum of deceptive strategies (e.g., embedded lies, omissions, denials), integrate cross-cultural perspectives (*56*), and focus on diverse contexts of deception (e.g., malicious intentions, self-presentations, personal opinions and feelings).

## Computational choices in automated verbal deception detection research

### *Text-embeddings are not necessarily better than simpler approaches*

Research in automated verbal deception detection has seen a methodological shift from *linguistic and statistical features* (i.e., features derived from dictionary-based methods, machine learning, or statistical properties of text such as LIWC, part-of-speech tags, named-entities, readability scores) toward representation learning approaches, such as *embeddings* (Fig. 3A; see also Table S5). That shift is particularly evident after 2015 when several landmark papers in NLP research were published (*27–30*). This transition reflects the evolution of NLP techniques for improved text representations and marks a substantial deviation from earlier practices, in which dictionary-based approaches dominated verbal credibility research (*25*).

Surprisingly, our meta-analytic findings showed that the shift from *linguistic and statistical features* to *embeddings representations* was associated only with negligible - albeit statistically significant – gains, increasing predicted accuracy by only 1.5 percentage points ($\beta = 0.12$, [0.11, 0.13], $p < .001$). However, this small average gain should be interpreted considering that the *embeddings* representation pools very different pipelines: embeddings fed to classical or neural classifiers, fine-tuned transformer-based models, and large language models (LLMs), which were mostly used through prompting. As detailed in the next section, transformer-based models, which almost exclusively rely on embeddings, outperformed statistical learning models by 3.5 percentage points, whereas LLMs performed worse. This pattern suggests that any advantage of embeddings may be masked by differences in model architecture and training procedure (e.g., fine-tuning vs. prompting). Because representation and architecture are confounded in the available data, their contributions cannot be fully separated.

---

[2] Other languages were underrepresented and constituted, altogether, 12.80% of reports (*n*=37; see the datafile in OSF for full details); eleven reports (3.81%) utilized mixed-language datasets, and in four reports (1.38%), the language of the datasets used was unspecified and unclear.

[3] The most frequently reused datasets were the Deceptive Opinion Spam datasets (*10*, *11*), which accounted for 41.34% of the reused datasets and were used in 25.61% (*n*=74) of all reports. However, these datasets contain important confounds related to data origin that may overestimate the reported models' performance (*94*), as genuine reviews were scraped from TripAdvisor, and fake reviews were crowdsourced from Amazon Mechanical Turk. The second most reused dataset was the Real-life trial dataset (*52*), a dataset of transcripts from videos of public court trial hearings, accounting for 13.41% of reused datasets and being used in 8.30% (*n*=24) of reports. Finally, the third most reused dataset entails customer-generated truthful reviews, crowdsourced-generated deceptive reviews, and employee-generated deceptive reviews (*95*), accounting for 11.73% of re-used datasets and used in only 7.27% (*n*=21) of reports.

Regardless of potential accuracy gains, employing embeddings comes at the expense of the interpretability that simpler, feature-based models offer. In our sample, only models based on *linguistic and statistical features* could be directly connected to theories[4]. Examples include previous studies employing such feature-based models to mimic manual approaches to credibility assessment, such as Reality Monitoring (*60–62*), or to operationalize common deception theories (*63–65*). Statistical caution for parsimony and practices in translational research in high-stakes contexts (e.g., court proceedings, airport security, fraud prevention) dictate that, whenever possible, simpler and more explainable models should be favored over opaque alternatives (*66*). In such scenarios, understanding the rationale behind algorithmic decisions is as crucial as the process underlying the decisions themselves (*66*); therefore, employing models grounded in psychological deception theories provides a significant advantage even if at the expense of reduced accuracy.

***Large language models do not outperform simpler approaches***

Statistical learning models remain the most frequently used for deception detection ($n$=2,973, 48.45%), but recent years have also seen the adoption of deep learning models, transformer architectures, and, after 2022 (*31*), the rise of LLMs (Fig. 3B; Table S5). One advantage of relying on pre-trained language models, such as BERT (*30*), FLAN-T5 (*32*), and Llama models (*67*), is the ability to fine-tune their pre-existing language representations for deception detection. Previous research that already adopted that approach showed how it reduces the need for a labor-intensive process of manually engineering features or determining which features to extract and select (*46*, *68*, *69*), which is a practice that often carries important consequences, such as variability in analytic decisions and fragmentation within research methodologies (*70*). By fine-tuning large language models, researchers can work directly with raw text inputs, streamlining workflows and enhancing consistency across studies.

However, our data indicate that the increasing adoption of more complex architectures does not uniformly translate to superior detection performance. Transformer-based architectures achieved the largest performance advantage over statistical learning models across model categories ($\beta = 0.32$, [0.30, 0.34], $p < .001$, 3.5 percentage points), whereas LLMs performed significantly worse ($\beta = -0.09$, [-0.13, -0.06], $p < .001$, -1.2 percentage points). This finding may reflect the distinction between task-specific fine-tuning and prompt-based approaches, as currently available LLMs are not optimized for deception detection tasks when used without additional adaptation. Therefore, the current evidence does not support the assumption that larger or more recent models will automatically yield more reliable deception-detection systems. Instead, future research should focus on identifying the conditions under which advanced architectures provide meaningful improvements.

[4] Among models relying on *linguistic and statistical features*, 47.21% ($n$=593) of models explicitly connected feature selection to theoretical frameworks or previous research, while 35.11% ($n$=441) of them were purely data-driven, and the remaining 17.68% ($n$=222) did not provide enough information.

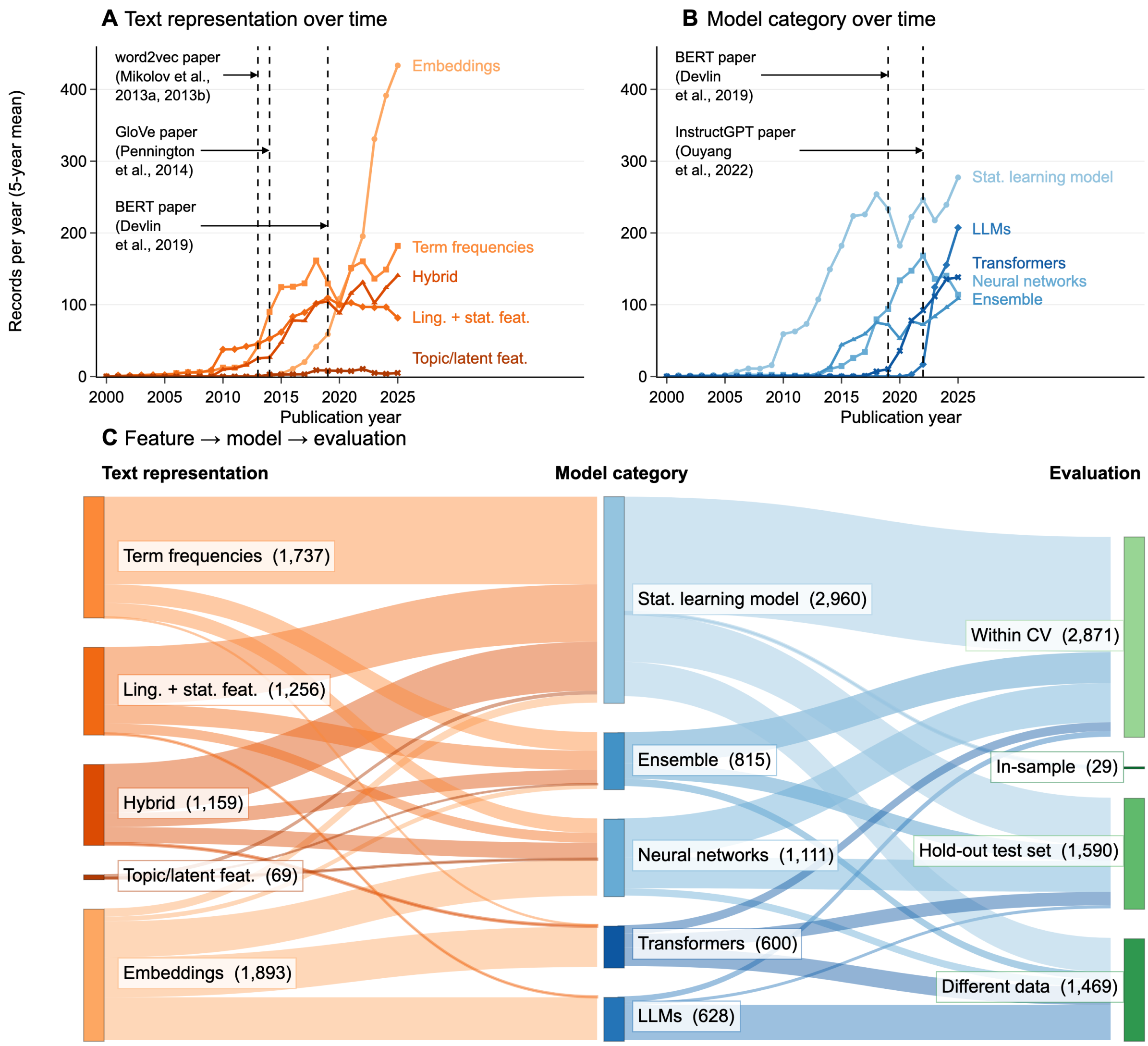

**Fig. 3 Text representations and model categories over time, and their interplay with evaluation procedures.**

(**A**) Text-representation approaches by publication year. (**B**) Model categories by publication year. (**C**) Sankey diagram illustrating the interplay between three methodological choices: text representation, model category, and evaluation procedure. In (A) and (B), one model from 1982, included in the systematic review, was excluded from the plot for a better visualization, and displayed values were computed using a 5-year rolling average to minimize volatility in the visualization; both panels share a common y-axis to allow direct comparison. Category names are abbreviated throughout: Ling. + stat. feat. = Linguistic and Statistical features; Topic/latent feat. = Topic and latent semantic features; Stat. learning model = Statistical Learning Model; Transformers = Transformer-based Models; LLMs = Large Language Models; Ensemble = Ensemble Models; Within CV = Within cross-validation. In (C), models coded as "Unclear" on a given dimension were excluded from the corresponding panel.

### *"Being accurate about accuracy"*[5]

Regarding evaluation procedures, in-sample validation - which is known to inflate accuracy estimates (*71*) - was used in only 0.47% of models (*n*=29), marking a striking deviation from psycholegal research practices, where in-sample procedures were commonly employed to evaluate manual approaches (*71*). In contrast, cross-validation was the most common practice in automated verbal deception detection research (*n*=2,871, 46.79% of models), followed by the use of a hold-out test set (*n*=1,590, 25.91%) (Fig. 3C; Table S5). Yet, only 23.96% of models were tested on an independent sample (*n*=1,470) (Fig. 3C). Therefore, adequate prediction stress testing remains limited: without testing on independent, unseen, or out-of-domain data, it is difficult to determine whether models can generalize beyond their training conditions (*57*, *72*). Previous studies have shown that deception detection performance can fall to chance level when applied across domains (i.e., a model trained to successfully detect deceptive opinions fails to detect deceptive intentions) (*69*, *73*) and cultures (i.e., some linguistic cues of deception translate across domains only when remain within the same language but do not transfer across different languages) (*56*). Our meta-analysis corroborates this: testing a model on new data reduces the predicted accuracy by up to 4.7 percentage points, on average, compared to within-dataset evaluation ($\beta$ = -0.38, [-0.40, -0.36], $p < .001$), whereas using a hold-out test set slightly inflates performance ($\beta$ = 0.06, [0.03, 0.09], $p < .001$, 0.6 percentage points). Future research should, therefore, consider testing models with demonstrated robust procedures using inner cross-validation and independent samples for testing to ensure fairer performance estimates (*72*, *74*).

### *Shared data, shared tasks, and common reporting practices*

More than half of the reports (*n*=162, 56.06%) have investigated deception detection by reanalyzing existing datasets (see Data S2), and only a minority collected new primary data (*n*=99, 34.26%)[6]. The practice of reanalyzing existing datasets and collecting new ones balances research replicability, as it facilitates the comparison of findings across studies (*75*), with the generation of new knowledge and resources to move the field forward. Nevertheless, formal reporting standards have yet to be established. To strengthen reproducibility and transparency, future work could adopt standardized practices, such as model cards (*76*) and pre-registration protocols (*77*). Model cards report key information about datasets, training procedures, evaluation methods, and contexts of use (*76*). Pre-registration of predictive modeling consists of pre-specifying details on i) research questions, ii) model training, and iii) evaluation procedures. Additionally, the pre-registration might involve "freezing" the trained model and subsequently testing it on newly collected data for evaluation. By adopting such practices, models and methodologies become more accessible and verifiable, reducing the current, scattered reporting practices. Beyond individual pre-registration, new avenues for making an impact in the field may include joining mass collaborative efforts to solve shared tasks (*78*, *79*). In a shared task framework, researchers compete to solve a shared problem under the same constraints and using the same dataset. By pooling independent analyses from different research teams, the predictability estimates become more credible and less likely to be dismissed due to concerns about overfitting or researcher degrees of freedom.

---

[5] This expression comes from Kleinberg et al. (2019), and has been borrowed to refer to the importance of using correct validation procedures to obtain fair estimates of performance. The results in this paragraph cite and take into consideration the results of Kleinberg et al. (2019).

[6] For 3.46% of reports (*n*=10), it was unclear whether they reused available datasets or collected a new one. Additionally, 6.23% of reports (*n*=18) used a mixed approach, combining newly collected data, previously available datasets, or data of unclear origin.

## Computational models remain the best available alternative

The apparent 70–75% performance ceiling indicates that the past 25 years of computational advances have not substantially increased the accuracy of automated verbal deception detection. Yet, this finding should be interpreted relative to the alternative approaches currently available. Unaided human judgments of deception are well-known to be consistently unreliable (*12*), even when originating from professionals who routinely assess credibility (*14*) and confidently believe in their deception detection ability (*14*, *80*). Some improvement is possible when judgments are constrained by empirically informed strategies. These include heuristic approaches that direct attention to a specific diagnostic feature, such as focusing on the statement's detailedness instead of veracity (*81*), approaches based on the occurrence of verifiable details (*38*, *39*), structured content-analysis frameworks, such as Criteria-Based Content Analysis and Reality Monitoring (*17*, *41*), and cognitive elicitation approaches (*16*).

Although these methods can improve performance relative to unaided judgment, they typically rely on intense manual coding, are difficult to implement at scale, and have often been validated on small sample sizes using in-sample evaluation (*71*). Against these approaches, we argue that computational methods for deception detection are a meaningful alternative and ought to be considered the state-of-the-art in verbal deception detection. In fact, computational models of deception can i) learn from multiple observations and features, enabling the integration of different theories and deception cues; ii) be retrained as new data become available, allowing for continuous learning and model updating; and iii) be employed at scale to generate deception predictions across large numbers of statements.

Yet, these advantages do not eliminate important limitations to their practical use. One concerns the tension between predictive performance and explainability. If more complex but opaque representations provide little or no performance advantage, simpler, theory-informed models should be preferred, particularly in high-stakes settings (*66*). At the same time, computational methods need not be restricted to fully automated deception judgments. A potentially more immediate application can be the automated coding of theoretically meaningful verbal cues, which could preserve a degree of interpretability while overcoming some of the scalability and resource limitations of manual coding. In parallel, future research should investigate whether advances in explainable AI can make embedding-based models more interpretable without sacrificing their potential predictive advantages.

A second limitation concerns how computational predictions of deception should be incorporated into human decision-making. Algorithmic judgments are often subjected to substantial criticism, whereas human judgments are frequently treated as the default alternative despite extensive evidence of their limited accuracy (*12–14*). Recent experimental evidence suggests that human involvement tends to reduce the performance of an otherwise accurate deception-detection model (*82*, *83*), while under other conditions, people may rely excessively on algorithmic advice (*84*). The relevant question for research and practice may therefore not be whether algorithms should replace humans, but how responsibilities should be allocated between humans and machines.

## Moving the needle of verbal deception research

With deception detection being a long-standing and unresolved task, moving beyond the methodological constraints of individual studies will require more than incremental refinement within current research conventions. We believe one of the field's biggest opportunities lies in moving from isolated datasets and domains to a unified deception corpus. Such a large-scale and multilingual corpus would support the adoption of mega-analyses and collaborative shared tasks in future research. Mega-analyses represent a valuable resource for systematically reanalyzing the most common verbal cues and deception theories (e.g., cognitive load, verifiability of details, reality monitoring) across different contexts, domains, and languages. Collaborative shared tasks ensure more robust findings, as independent analyses are pooled from different research teams. Together, these research practices would strengthen the available theoretical framework on verbal deception.

Furthermore, rather than repeatedly fine-tuning new architectures on new datasets for each domain, future research should consider fine-tuning language models across the full breadth of available deception corpora to ultimately develop a foundation model of deception: a large-scale model trained on broad data that can be adapted to a wide range of downstream tasks (*85*). If trained effectively, such a model could learn deception-related cues that generalize across multiple languages, deception forms, and topics, rather than relying on the idiosyncrasies of any single dataset. Its real value would lie in providing future research with a shared starting point to fine-tune from. Similar to what was argued on foundation models of cognition (*86*), we believe a foundation model of deception is an important advancement that can move the field from a collection of contextual experimentations to a cumulative scientific endeavor.

## Conclusion

The present work systematically analyzed 25 years of research on automated verbal deception detection, mapping research practices across 289 reports and 6,136 models and meta-analyzing the performance of 3,653 models. Across studies, machine learning models showed a pooled accuracy of 74.4% (95% CI: 71.2%-77.4%), closely aligning with previous meta-analytic estimates from different verbal deception detection paradigms. Together the evidence converges on a ceiling of 70%-75%. The pooled performance was significantly moderated by methodological choices: i) ground truth was clearly specified in only 12.46% of reports; ii) deception research has remained relatively narrow in scope, focusing primarily on fabrication (63.67% of reports), English-language datasets (82.01%), and tasks involving fake reviews (41.52% of reports) and past experiences (15.92%); iii) an unresolved tension remains between ecologically valid settings and high internal validity. Over time, the field has adopted more complex computational techniques, including text embeddings and neural model architectures. However, the apparent methodological progress did not translate into improved predicted performance. Further, only a few models (23.96%) were tested on new data to assess out-of-sample performance. Despite these limitations, computational models of deception already outperform unaided human judgment and match manual methods in accuracy while surpassing them in scalability and reliability. Advancing the field further requires more than methodological refinement within existing paradigms: progress demands a shift toward large-scale and cross-domain approaches, multi-lab collaborative efforts, and ultimately the development of a foundation model of deception.

## Materials and Methods

Data, code, and materials availability
The coded spreadsheet of included reports used in this systematic review and meta-analysis, together with the code used to analyze the data, is openly available via the Open Science Framework (OSF) at: https://osf.io/rmuq8/overview?view_only=dda72637ee604c94bf72135496e5f4de

Protocol and Preregistration
This work followed the Preferred Reporting Items for Systematic reviews and Meta-Analyses (PRISMA) protocol (*87*). The systematic review protocol was preregistered following the Generalized Systematic Review Registration in Open Science Framework (OSF) at: https://osf.io/wxme6/overview?view_only=e5308e8be0d04476ace9c102c1481489.

Eligibility criteria
Identified reports were selected according to predefined inclusion and exclusion criteria. First, eligible reports were empirical works, published in English in peer-reviewed journals, conference proceedings, book chapters, and preprints. Opinion papers, systematic reviews, and meta-analyses were excluded. Second, reports eligible for this review had to address the topic of automated detection of human deceptive statements (either typed or transcribed) from verbal cues. Following our definition (see Main text), only reports that applied automated techniques for both the extraction of textual features or verbal cues and for the prediction judgment were included: all types of verbal cues were considered eligible, as long as they were automatically extracted. In contrast, research targeting non-verbal indicators, such as vocal, facial, body, or physiological signals, was omitted. Reports on multimodal deception detection were included if they also reported the performance of models trained exclusively on textual features. Third, eligible reports had to utilize predictive models and report classification metrics in their analysis; research solely reporting explanatory findings (e.g., using ANOVA for explaining main and interaction effects of factors), relying on unsupervised methods (e.g., clustering), or focusing on the accuracy of human judgment - in the absence of a computational model - was excluded. Finally, all forms of human deception (e.g., fabrication, omissions, embedded lies) were eligible, but research related to i) fake news or misinformation, ii) deception by LLMs, and iii) the detection of human-written vs AI-generated content was excluded, as - while related to deception - it constitutes separate research domains.

Search strategy
Seven databases were searched for relevant research. Literature in computer science and computational linguistics was searched in (1) the *ACL Anthology* (on March 21st 2025), (2) the *ACM Digital Library* (on March 5th 2025), (3) *IEEEXplore* (on March 12th 2025). Literature in psychology was searched in (4) *Web of Science* (on March 5th 2025), (5) *Scopus* (on March 5th 2025), and (6) *PsycINFO* (on March 16th 2025). Preprints were searched on (7) *ArXiv* (on March 13th 2025). The search strategy included terms associated with the following semantic areas: (1) automated, (2) verbal, (3) deception, and (4) detection (see the Search string section for the full search string). A total of 23,773 records were identified. After removing 7,520 duplicates, 153 books, 147 dissertations, 13 retracted papers, and 132 records marked ineligible by Zotero, 15,808 records remained for title-and-abstract screening. We confirmed the quality of our search strategy by double-checking the inclusion of published key reports from two past relevant reviews and meta-analyses on deception detection (*25*, *33*). Key reports were defined and preregistered before starting the search (Data S1) and used later to also assess the quality of the title-and-abstract screening phase.

Search string
The employed full search string is reported below. In bold the main keyword, followed by the other semantically-related terms.

((**automat*** OR comput* OR "machine learning" OR "deep learning" OR "natural language processing" OR AI OR "artificial intelligence" OR "language models" OR LLMs)
AND
(**verbal** OR text* OR narrative* OR written OR statement* OR content)
AND
(**decept*** OR lie* OR lying OR deceit* OR dishonest* OR credibility OR veracity OR truth* OR believability)
AND
(**detect*** OR identif*))

NOT
(“fake news” OR disinformation OR misinformation).

Title-and-abstract screening with ASReview LAB

Title-and-abstract screening (until March 2025) was conducted using ASReview LAB v.1 (https://asreview.nl) and resulted in 415 reports deemed eligible for retrieval. ASReview LAB is a free, open-source tool for screening and labeling a large collection of reports for systematic reviews and meta-analysis (*88*). The core of the tool relies on active learning, wherein the researcher – as human-in-the-loop - labels records as relevant or irrelevant for the review and is in interaction with a machine learning (ML) model, which updates a ranking algorithm to continuously re-rank the literature reports from the most to the least relevant. Each time the researcher labels a new record, the rank order is updated. The active learning cycle is repeated until a stopping rule is reached (i.e., a predefined criterion that makes the screeners sufficiently confident that all relevant records were seen). For the initial search of this review, title-and-abstract screening via ASReview was conducted by two screeners, who independently reviewed records, following the SAFE procedure detailed below *(89)*.

1. *Screening a random set for training data:*

   The screening with AS Review requires adding prior knowledge to a model (i.e., a small set of reports labeled as relevant or irrelevant). The best procedure recommended in the SAFE guidelines *(89)* consisted of labeling a random and stratified 1% of total records (*n*=158 out of 15,808) for the training data. Stratification was done for years of publication, database (e.g., ACL, ArXiv, PsycInfo), and type of document (e.g., preprint, full paper, conference proceeding). Two researchers independently screened the records from the same training set and solved disagreements by discussion. Of the 158 records, eight were marked as relevant and 150 as irrelevant.

2. *Title-and-abstract screening with active learning:*

   Two screeners uploaded the same training data labeled in Step 1 as prior knowledge to train the same active learning model, but started screening records independently. For this stage, a naïve-bayes model was trained on a TF-IDF document representation.

   Both screeners stopped the screening after all of the following criteria were met *(89)*:
   - All key records were marked as *relevant*;
   - At least 10% of the total records were screened (*n*=1,581);
   - At least twice the estimate of relevant records (ERR)[7] was screened (*n*=1,600);
   - No relevant records were identified in the last 50 consecutive records.

   After the independent screening ended, the records that were seen by only one of the two screeners were evaluated by the other screener, and all disagreements were resolved by discussion. A total of 1,693 titles and abstracts were screened, of which 381 (22.50%) were labeled as relevant.

3. *Locating hard-to-find reports:*

   As a safeguard against missing records in the screening procedure described above, a final round of automated screening was performed by switching to a more advanced text classification model (see Teijema et al., 2023). With the partly labeled dataset obtained from Step 2 as prior knowledge, we used an XGBoost model trained on a TF-IDF[8] text representation to initiate a new active-learning cycle. The stopping rule was set to 100 consecutive irrelevant records. After the independent screening ended, records seen by only one of the two screeners were evaluated by the other screener, and all disagreements were resolved by discussion. In this phase, an additional 375 titles and abstracts were screened, with six records identified as relevant.

[7] The ERR is a crude estimate of the number of relevant records in the total dataset computed by dividing the number of relevant records found in the training set by the total number of records in the training set, multiplied by the total number of records in the whole dataset. In this case, the estimate of relevant reports was equal to ERR = (8/158) * 15,808 = 800.

[8] We preregistered this phase slightly differently as we aimed to employ a logistic regression trained on a sBERT document representation to rely on embeddings instead of word-frequency. However, given that the switch to a transformer-based model was too computationally intense for our resources, we decided to keep the TF-IDF text representation and switch to a more computationally complex model for training (here: XGBoost instead of logistic regression). The stopping rule was initially set to 50 irrelevant consecutive reports but then increased to 100 for a more extensive screening.

4. *Quality check for excluded reports:*

Lastly, to assess whether records were incorrectly excluded (e.g., due to screening fatigue), a second screening was conducted on excluded records (*89*). This quality check was run by using as prior knowledge the ten highest- and lowest-ranked reports from Phase 3 with a simple model (here: naïve bayes + TF-IDF). In other words, the two screeners double-checked for the initially excluded records but rank-ordered on relevance scores until the stopping rule of 50 consecutive excluded records was reached. After the independent screening ended, records seen by only one of the two screeners were evaluated by the other screener, and disagreements were solved by discussion. Of 1,831 initially excluded records, 220 title-and-abstracts were double-checked, and 20 reports were marked back as relevant.

Update search
The initial search covered records published until March 2025. To identify studies published during the remainder of 2025, one reviewer reran the same search query in January 2026, restricting the search to records indexed or published in 2025. The update search was run on all databases from the previous search round except for *ACM Digital Library*, which has been updated at the beginning of January 2026 and led to an unrealistic number of records that even exceeded the number found in the previous search, suggesting a change in the database search reliability. This updated search yielded 5,250 records, of which 739 were duplicates, and 11 were marked as ineligible by Zotero, leaving 4,500 additional records for title-and-abstract screening. The title-and-abstract screening was conducted by a single reviewer using ASReview. The active learning model was a naïve-bayes classifier trained on TF-IDF document representations, using records labeled as relevant or irrelevant during the initial screening phase as prior knowledge for the updated screening. The title-and-abstract screening stopped after 100 consecutive irrelevant records were found. In total, 405 records were screened, of which 164 were marked as eligible for full-text screening.

Full-text screening
After the initial and updated titles-and-abstracts screening, 579 eligible reports were assessed for retrieval and full-text screening. Of these, eleven could not be retrieved, and 71 were duplicates, leaving 497 reports for full-text eligibility assessment. After full-text review, 208 reports were excluded for reasons such as: language other than English ($n$=1), focus on human deception detection ($n$=3), use of synthetic/generated data ($n$=19), non-empirical work ($n$=11), manual extraction of verbal cues ($n$=5), low quality of reporting ($n$=28), focus on fake news/rumor detection ($n$=28), no use of supervised ML models ($n$=33), and lack of focus on verbal deception detection ($n$=80). Ultimately, 289 reports were included in the review: 136 (47.06%) peer-reviewed journal articles, 135 (46.71%) conference papers, and 18 (6.23%) preprints. Full list of included reports is available at https://osf.io/rmuq8/overview?view_only=dda72637ee604c94bf72135496e5f4de.

Coding scheme
Data extraction was conducted on the full-text versions of the final set of 289 reports, using the coding scheme available at https://osf.io/rmuq8/overview?view_only=dda72637ee604c94bf72135496e5f4de. We piloted the initial coding scheme and trained three reviewers by independently coding 42 reports (16.93% of reports from the initial search), which allowed us to refine the coding scheme to its final form and confirm inter-coder agreement on the included reports (percentage agreement = 0.78-0.79). Based on a satisfactory agreement and the large number of reports, we assigned a subset of reports to three independent reviewers[9]. All reviewers had the possibility to discuss with each other how to handle difficult cases.

Full texts were coded at the model-level, meaning we created a separate entry for each model trained to detect automated verbal deception, with multiple entries within the same report. This approach allowed us to capture fine-grained information at the model-level while maintaining the ability to aggregate results at the report level when appropriate. Following an inductive (bottom-up) approach, variables, such as type of deception and topic of investigation, dataset reuse, language, and size, text-representation, model categories, and evaluation approach, were first coded following what was reported in the full-text and then eventually mapped into overarching categories. In contrast, a deductive (top-down) approach, in which full texts were coded using pre-defined codes, was employed for coding the variables of the data source and ground truth. The glossary in Table 1 (see Main text) reports definitions and examples for each level of these overarching variables.

[9] Initially, we planned to have two independent coders coding full texts of all included studies, followed by verification through inter-rater agreement and discussion. However, given the large number of included reports, we decided to split the task between three coders, after reaching substantial agreement on an initial training set of 42 reports.

Meta-analytical plan

To estimate pooled classification performance for verbal deception detection, a multilevel meta-analysis was conducted. The unit of analysis was each individual model's performance metrics reported in the reports. Among all reported metrics, accuracy was the most frequently reported (75.7 % of models) and could be expressed as the ratio of correct predictions out of the total set of predictions, allowing for a meta-analysis of proportions. The test set size was determined according to the evaluation protocol: a) for models evaluated using a hold-out procedure, test set size was computed from the reported train–test split; b) for models evaluated using within cross-validation, the test set size was defined as the total dataset size, as all observations contribute to the cross-validation procedure; c) for models evaluated on an independent dataset, the test set size corresponded to the size of the external evaluation dataset reported in the original report. Models that did not report classification accuracy ($n$=1,491), evaluated deception using non-standard truthful vs. deception contrasts (e.g., deception vs. deception; $n$=123), or lacked sufficient information (i.e., contained levels equal to "*No info*" or "*Unclear*") regarding the data source, ground-truth, text-representation, model category, or evaluation procedure ($n$ = 328) were excluded from this quantitative synthesis. All analyses were conducted in R using the *metafor* package (*91*).

Once extracted the number of correct predictions and test-set size for each included model, effect sizes (i.e., logit-transformed proportions; *PLO*) and their sampling variances were computed with the *escalc* function. Because several reports could contribute multiple classification models, and multiple models could be evaluated on the same dataset, dependencies among effect sizes were accounted for by specifying a random intercept at both the dataset and report levels. Additionally, as classification accuracy is strongly influenced by dataset class imbalance, the proportion of the majority class was included as a continuous moderator. This variable was centered at 0.50 such that the model intercept corresponded to the expected classification accuracy under balanced class distributions. The equation of the first multilevel meta-regression was the following:

$$Accuracy \sim majority\ class + (1 \mid record_id) + (1 \mid dataset_id)$$

Moderator analysis of methodological choices was performed using a multilevel meta-regression model with the following equation:

$$\begin{gathered} Accuracy \sim source + ground\ truth + text\ representation + model\ category \\ + evaluation\ procedure + majority\ class \\ + (1 \mid record_id) + (1 \mid dataset_id) \end{gathered}$$

Treatment contrasts were used throughout, with reference categories defined a priori based on conventional or theoretically meaningful baselines. Omnibus Wald tests were conducted to evaluate the overall contribution of each categorical moderator while controlling for the remaining variables in the model. Heterogeneity in the multilevel models was assessed with Cochran's Q test for residual heterogeneity (QE), the REML variance components at the dataset and report levels ($\sigma^2$, with 95% profile-likelihood CIs), and multilevel $I^2$, computed with the matrix-based method of the *i2_ml*() function in the *orchaRd* package. Variance explained by the models was quantified separately with marginal and conditional $R^2$ from the *r2_ml*() function in the *orchaRd* package. Unless otherwise specified, statistical significance was assessed using two-sided tests with $\alpha$ = .05. Regression coefficients are reported on the logit scale, whereas predicted accuracies are presented on the original proportion scale together with their corresponding 95% confidence intervals.

Publication bias was not statistically assessed because ML models' performance was typically not accompanied by a $p$-value, making traditional publication bias analyses unfeasible in this context. Nevertheless, publication bias was mitigated by including preprints during the screening phase and by coding all reported models (rather than only the best-performing ones) in each report during the data extraction phase.

## References

1. N. Abe, The neurobiology of deception: Evidence from neuroimaging and loss-of-function studies. *Curr. Opin. Neurol.* **22**, 594–600 (2009).
2. K. B. Serota, T. R. Levine, A Few Prolific Liars. *J. Lang. Soc. Psychol.* **34**, 138–157 (2015).
3. B. M. DePaulo, S. E. Kirkendol, D. A. Kashy, M. M. Wyer, J. A. Epstein, Lying in Everyday Life. *J. Pers. Soc. Psychol.* **70**, 979–995 (1996).
4. R. Halevy, S. Shalvi, B. Verschuere, Being Honest about Dishonesty: Correlating Self-Reports and Actual Lying. *Hum. Commun. Res.* **40**, 54–72 (2014).
5. B. L. Verigin, E. H. Meijer, G. Bogaard, A. Vrij, Lie prevalence, lie characteristics and strategies of self-reported good liars. *PLoS One* **14** (2019).
6. T. Brennen, S. Magnussen, Lie Detection: What Works? *Curr. Dir. Psychol. Sci.* **32**, 395–401 (2023).
7. J. Morgan, Wrongful convictions and claims of false or misleading forensic evidence. *J. Forensic Sci.* **68**, 908–961 (2023).
8. S. M. Kassin, S. A. Drizin, T. Grisso, G. H. Gudjonsson, R. A. Leo, A. D. Redlich, Police-induced confessions: Risk factors and recommendations. *Law Hum. Behav.* **34**, 3–38 (2010).
9. S. L. Humpherys, K. C. Moffitt, M. B. Burns, J. K. Burgoon, W. F. Felix, Identification of fraudulent financial statements using linguistic credibility analysis. *Decis. Support Syst.* **50**, 585–594 (2011).
10. M. Ott, C. Cardie, J. T. Hancock, Negative Deceptive Opinion Spam. Association for Computational Linguistics [Preprint] (2013). https://aclanthology.org/N13-1053/.
11. M. Ott, Y. Choi, C. Cardie, J. T. Hancock, Finding deceptive opinion spam by any stretch of the imagination. *ACL-HLT 2011 - Proceedings of the 49th Annual Meeting of the Association for Computational Linguistics: Human Language Technologies* **1**, 309–319 (2011).
12. C. F. Bond, B. M. DePaulo, Accuracy of deception judgments. *Personal. Soc. Psychol. Rev.* **10**, 214–234 (2006).
13. M. Hartwig, C. F. Bond, Why do lie-catchers fail? A lens model meta-analysis of human lie judgments. *Psychol. Bull.* **137**, 643–659 (2011).
14. M. Aamodt, H. Custer, Who can best catch a liar? *Forensic Examiner* **15(1)**, 6 (2006).
15. M. Hartwig, P. A. Granhag, T. Luke, Strategic Use of Evidence During Investigative Interviews: The State of the Science. *Credibility Assessment: Scientific Research and Applications*, 1–36 (2014).
16. A. Vrij, R. P. Fisher, H. Blank, A cognitive approach to lie detection: A meta-analysis. *Legal Criminol. Psychol.* **22**, 1–21 (2015).
17. Y. Gancedo, F. Fariña, D. Seijo, M. Vilariño, R. Arce, Reality Monitoring: A Meta-analytical Review for Forensic Practice. *European Journal of Psychology Applied to Legal Context* **13**, 99–110 (2021).
18. B. G. Amado, R. Arce, F. Fariña, Undeutsch hypothesis and Criteria Based Content Analysis: A meta-analytic review. *European Journal of Psychology Applied to Legal Context* **7**, 3–12 (2015).
19. B. G. Amado, R. Arce, F. Fariña, M. Vilariño, Criteria-Based Content Analysis (CBCA) reality criteria in adults: A meta-analytic review. *International Journal of Clinical and Health Psychology* **16**, 201–210 (2016).
20. S. L. Sporer, V. Hauch, J. Masip, N. Martschuk, A Meta-Analysis of Field Studies on Criteria-Based Content Analysis. *Eur. Psychol.*, doi: 10.1027/1016-9040/A000561 (2025).
21. Y. Tausczik, J. P.-J. of language and, undefined 2010, The psychological meaning of words: LIWC and computerized text analysis methods. *journals.sagepub.com* **29**, 24–54 (2010).
22. M. Ali, T. Levine, The Language of Truthful and Deceptive Denials and Confessions. *Communication Reports* **21**, 82–91 (2008).
23. G. D. Bond, A. Y. Lee, Language of lies in prison: linguistic classification of prisoners' truthful and deceptive natural language. *Appl. Cogn. Psychol.* **19**, 313–329 (2005).
24. M. L. Newman, J. W. Pennebaker, D. S. Berry, J. M. Richards, Lying words: Predicting deception from linguistic styles. *Pers. Soc. Psychol. Bull.* **29**, 665–675 (2003).
25. V. Hauch, I. Blandón-Gitlin, J. Masip, S. L. Sporer, Are Computers Effective Lie Detectors? A Meta-Analysis of Linguistic Cues to Deception. *Personality and Social Psychology Review* **19**, 307–342 (2015).
26. Y. Lecun, Y. Bengio, G. Hinton, Deep learning. *Nature 2015 521:7553* **521**, 436–444 (2015).
27. T. Mikolov, K. Chen, G. Corrado, J. Dean, Efficient Estimation of Word Representations in Vector Space. *1st International Conference on Learning Representations, ICLR 2013 - Workshop Track Proceedings* (2013).
28. T. Mikolov, I. Sutskever, K. Chen, G. S. Corrado, J. Dean, Distributed Representations of Words and Phrases and their Compositionality. *Adv. Neural Inf. Process. Syst.* **26** (2013).

29. J. Pennington, R. Socher, C. D. Manning, GloVe: Global Vectors for Word Representation. *Proceedings of the 2014 Conference on Empirical Methods in Natural Language Processing (EMNLP)*, 1532–1543 (2014).
30. J. Devlin, M. W. Chang, K. Lee, K. Toutanova, BERT: Pre-training of Deep Bidirectional Transformers for Language Understanding. *NAACL HLT 2019 - 2019 Conference of the North American Chapter of the Association for Computational Linguistics: Human Language Technologies - Proceedings of the Conference* **1**, 4171–4186 (2019).
31. L. Ouyang, J. Wu, X. Jiang, D. Almeida, C. L. Wainwright, P. Mishkin, C. Zhang, S. Agarwal, K. Slama, A. Ray, J. Schulman, J. Hilton, F. Kelton, L. Miller, M. Simens, A. Askell, P. Welinder, P. Christiano, J. Leike, R. Lowe, Training language models to follow instructions with human feedback. *Adv. Neural Inf. Process. Syst.* **35** (2022).
32. H. W. Chung, L. Hou, S. Longpre, B. Zoph, Y. Tay, W. Fedus, Y. Li, X. Wang, M. Dehghani, S. Brahma, A. Webson, S. S. Gu, Z. Dai, M. Suzgun, X. Chen, A. Chowdhery, A. Castro-Ros, M. Pellat, K. Robinson, D. Valter, S. Narang, G. Mishra, A. Yu, V. Zhao, Y. Huang, A. Dai, H. Yu, S. Petrov, E. H. Chi, J. Dean, J. Devlin, A. Roberts, D. Zhou, Q. V. Le, J. Wei, Scaling Instruction-Finetuned Language Models. (2022).
33. A. S. Constancio, D. F. Tsunoda, H. de Fátima Nunes Silva, J. M. da Silveira, D. R. Carvalho, Deception detection with machine learning: A systematic review and statistical analysis. *PLoS One* **18** (2023).
34. D. M. Markowitz, "Verbal and Linguistic Analysis of Deceptive Statements: A 20-Year Explication (2005-2025)" in *OSF Preprint* (2026; https://doi.org/10.31234/osf.io/3he2a_v1).
35. M. J. Salganik, *Bit by Bit: Social Research in the Digital Age* (Princeton University Press., 2019; https://books.google.com/books?hl=en&lr=&id=58iXDwAAQBAJ&oi=fnd&pg=PR1&dq=Salganik,+M.+J.+(2019).+Bit+by+bit:+Social+research+in+the+digital+age.+Princeton+University+Press.+ISBN:+978-0-691-19610-7&ots=0SpWv4e2ac&sig=G-EwTH_2cqZ4xrmV8WD9ONwns6M).
36. R. L. Boyd, A. Ashokkumar, S. Seraj, J. W. Pennebaker, The Development and Psychometric Properties of LIWC-22. (2022).
37. J. W. Pennebaker, R. J. Botth, R. L. Boyd, M. E. Francis, "Linguistic Inquiry and Word Count: LIWC2015" in *Austin, TX: Pennebaker Conglomerates* (2015).
38. N. Palena, L. Caso, A. Vrij, G. Nahari, The Verifiability Approach: A Meta-Analysis. *J. Appl. Res. Mem. Cogn.* **10**, 155–166 (2021).
39. B. Verschuere, G. Bogaard, E. Meijer, Discriminating deceptive from truthful statements using the verifiability approach: A meta-analysis. *Appl. Cogn. Psychol.* **35**, 374–384 (2021).
40. M. Hartwig, C. F. Bond, Lie detection from multiple cues: A meta-analysis. *Appl. Cogn. Psychol.* **28**, 661–676 (2014).
41. V. A. Oberlader, C. Naefgen, J. Koppehele-Gossel, L. Quinten, R. Banse, A. F. Schmidt, Validity of content-based techniques to distinguish true and fabricated statements: A meta-analysis. *Law Hum. Behav.* **40**, 440–457 (2016).
42. R. M. Hoffer, The importance of ground truth data in remote sensing. [Preprint] (1972).
43. S. Kühne, R. Aachen, G. B. Paul, Gut Feelings and Algorithms: Searching for Harmful Intentions in Airport Security Processes. *Engag. Sci. Technol. Soc.* **10**, 120–146–120–146 (2024).
44. A. Vrij, P. A. Granhag, S. Porter, Pitfalls and opportunities in nonverbal and verbal lie detection. *Psychological Science in the Public Interest, Supplement* **11**, 89–121 (2010).
45. D. M. Markowitz, Deconstructing deception: Frequency, communicator characteristics, and linguistic features of embeddedness. *Appl. Cogn. Psychol.* **38**, e4215 (2024).
46. R. Loconte, B. Kleinberg, Examining embedded lies through computational text analysis. *Scientific Reports 2025 15:1* **15**, 26482- (2025).
47. M. M. B. Alsaad, H. Joshi, A Framework for Online Fake Review Detection on Yelp Electronics Product Dataset Using Machine Learning Techniques. *Lecture Notes in Networks and Systems* **1305 LNNS**, 409–423 (2025).
48. L. Samineni, A. Peddi, A. Kasukurthi, M. V. P. C. S. Rao, G. Niharika, S. V. Chereddy, Evaluation of AI Techniques for Detecting Deceptive Reviews in Cyberspace: A Study of Pre- and Post-COVID-19 Trends. *Proceedings of the 2023 2nd International Conference on Electronics and Renewable Systems, ICEARS 2023*, 961–967 (2023).
49. D. A. Boumber, F. Z. Qachfar, R. Verma, Domain-Agnostic Adapter Architecture for Deception Detection: Extensive Evaluations with the DIFrauD Benchmark. [Preprint] (2024). https://aclanthology.org/2024.lrec-main.468/.
50. T. Bhatia, J. Meena, Detection of Fake Online Recruitment Using Machine Learning Techniques. *Proceedings - 2022 4th International Conference on Advances in Computing, Communication Control and Networking, ICAC3N 2022*, 300–304 (2022).

51. P. Lohumi, K. Gupta, V. Sharma, R. Jain, Enhancing the Detection of Deceptive Job Advertisements: A Machine Learning Approach with Advanced Oversampling and Vectorization Techniques. *Lecture Notes in Networks and Systems* **1270 LNNS**, 105–122 (2025).
52. V. Pérez-Rosas, M. Abouelenien, R. Mihalcea, M. Burzo, Deception detection using real-life trial data. *ICMI 2015 - Proceedings of the 2015 ACM International Conference on Multimodal Interaction*, 59–66 (2015).
53. T. R. Levine, Ecological Validity and Deception Detection Research Design. *Commun. Methods Meas.* **12**, 45–54 (2018).
54. T. D. Cook, D. T. Campbell, W. Shadish, *Experimental and Quasi-Experimental Designs for Generalized Causal Inference* (Boston, MA: Houghton Mifflin., 2002)vol. 1195.
55. J. P. Blair, T. R. Levine, A. S. Shaw, Content in Context Improves Deception Detection Accuracy. *Hum. Commun. Res.* **36**, 423–442 (2010).
56. K. Papantoniou, P. Papadakos, T. Patkos, G. Flouris, I. Androutsopoulos, D. Plexousakis, Deception detection in text and its relation to the cultural dimension of individualism/collectivism. *Nat. Lang. Eng.* **28**, 545–606 (2022).
57. T. Yarkoni, J. Westfall, Choosing Prediction Over Explanation in Psychology: Lessons From Machine Learning. *Perspect. Psychol. Sci.* **12**, 1100–1122 (2017).
58. J. M. Hofman, D. J. Watts, S. Athey, F. Garip, T. L. Griffiths, J. Kleinberg, H. Margetts, S. Mullainathan, M. J. Salganik, S. Vazire, A. Vespignani, T. Yarkoni, Integrating explanation and prediction in computational social science. *Nature 2021 595:7866* **595**, 181–188 (2021).
59. R. L. Boyd, H. A. Schwartz, Natural Language Analysis and the Psychology of Verbal Behavior: The Past, Present, and Future States of the Field. *J. Lang. Soc. Psychol.* **40**, 21–41 (2021).
60. B. Kleinberg, I. van der Vegt, A. Arntz, B. Verschuere, Detecting deceptive communication through linguistic concreteness. doi: 10.31234/OSF.IO/P3QJH (2019).
61. R. Loconte, C. Battaglini, S. Maldera, P. Pietrini, G. Sartori, N. Navarin, M. Monaro, Detecting Deception Through Linguistic Cues: From Reality Monitoring to Natural Language Processing. *J. Lang. Soc. Psychol.*, doi: 10.1177/0261927X251316883/SUPPL_FILE/SJ-DOCX-1-JLS-10.1177_0261927X251316883.DOCX (2025).
62. M. Schutte, G. Bogaard, E. Mac Giolla, L. Warmelink, B. Kleinberg, B. Verschuere, Man versus Machine: Comparing manual with LIWC coding of perceptual and contextual details for verbal lie detection. doi: 10.31234/OSF.IO/CTH58 (2021).
63. J. Sarzynska-Wawer, A. Pawlak, J. Szymanowska, K. Hanusz, A. Wawer, Truth or lie: Exploring the language of deception. *PLoS One* **18**, e0281179 (2023).
64. B. Kleinberg, M. Mozes, A. Arntz, B. Verschuere, Using named entities for computer-automated verbal deception detection. *J. Forensic Sci.* **63**, 714–723 (2017).
65. B. Kleinberg, Y. van der Toolen, A. Vrij, A. Arntz, B. Verschuere, Automated verbal credibility assessment of intentions: The model statement technique and predictive modeling. *Appl. Cognit. Psychol.* **32**, 354–366 (2018).
66. M. Oswald, J. Grace, S. Urwin, G. C. Barnes, Algorithmic risk assessment policing models: lessons from the Durham HART model and 'Experimental' proportionality. *Information & Communications Technology Law* **27**, 223–250 (2018).
67. A. Grattafiori, A. Dubey, A. Jauhri, A. Pandey, A. Kadian, A. Al-Dahle, A. Letman, A. Mathur, A. Schelten, A. Vaughan, A. Yang, A. Fan, A. Goyal, A. Hartshorn, A. Yang, A. Mitra, A. Sravankumar, A. Korenev, A. Hinsvark, A. Rao, A. Zhang, A. Rodriguez, A. Gregerson, A. Spataru, B. Roziere, B. Biron, B. Tang, B. Chern, C. Caucheteux, C. Nayak, C. Bi, C. Marra, C. McConnell, C. Keller, C. Touret, C. Wu, C. Wong, C. C. Ferrer, C. Nikolaidis, D. Allonsius, D. Song, D. Pintz, D. Livshits, D. Wyatt, D. Esiobu, D. Choudhary, D. Mahajan, D. Garcia-Olano, D. Perino, D. Hupkes, E. Lakomkin, E. AlBadawy, E. Lobanova, E. Dinan, E. M. Smith, F. Radenovic, F. Guzmán, F. Zhang, G. Synnaeve, G. Lee, G. L. Anderson, G. Thattai, G. Nail, G. Mialon, G. Pang, G. Cucurell, H. Nguyen, H. Korevaar, H. Xu, H. Touvron, I. Zarov, I. A. Ibarra, I. Kloumann, I. Misra, I. Evtimov, J. Zhang, J. Copet, J. Lee, J. Geffert, J. Vranes, J. Park, J. Mahadeokar, J. Shah, J. van der Linde, J. Billock, J. Hong, J. Lee, J. Fu, J. Chi, J. Huang, J. Liu, J. Wang, J. Yu, J. Bitton, J. Spisak, J. Park, J. Rocca, J. Johnstun, J. Saxe, J. Jia, K. V. Alwala, K. Prasad, K. Upasani, K. Plawiak, K. Li, K. Heafield, K. Stone, K. El-Arini, K. Iyer, K. Malik, K. Chiu, K. Bhalla, K. Lakhotia, L. Rantala-Yeary, L. van der Maaten, L. Chen, L. Tan, L. Jenkins, L. Martin, L. Madaan, L. Malo, L. Blecher, L. Landzaat, L. de Oliveira, M. Muzzi, M. Pasupuleti, M. Singh, M. Paluri, M. Kardas, M. Tsimpoukelli, M. Oldham, M. Rita, M. Pavlova, M. Kambadur, M. Lewis, M. Si, M. K. Singh, M. Hassan, N. Goyal, N. Torabi, N. Bashlykov, N. Bogoychev, N. Chatterji, N. Zhang, O. Duchenne, O. Çelebi, P. Alrassy, P. Zhang, P. Li, P. Vasic, P. Weng, P. Bhargava, P. Dubal, P. Krishnan, P. S. Koura, P. Xu, Q. He, Q. Dong, R. Srinivasan, R. Ganapathy, R. Calderer, R. S. Cabral, R. Stojnic, R. Raileanu, R. Maheswari, R. Girdhar, R. Patel, R.

Sauvestre, R. Polidoro, R. Sumbaly, R. Taylor, R. Silva, R. Hou, R. Wang, S. Hosseini, S. Chennabasappa, S. Singh, S. Bell, S. S. Kim, S. Edunov, S. Nie, S. Narang, S. Raparthy, S. Shen, S. Wan, S. Bhosale, S. Zhang, S. Vandenhende, S. Batra, S. Whitman, S. Sootla, S. Collot, S. Gururangan, S. Borodinsky, T. Herman, T. Fowler, T. Sheasha, T. Georgiou, T. Scialom, T. Speckbacher, T. Mihaylov, T. Xiao, U. Karn, V. Goswami, V. Gupta, V. Ramanathan, V. Kerkez, V. Gonguet, V. Do, V. Vogeti, V. Albiero, V. Petrovic, W. Chu, W. Xiong, W. Fu, W. Meers, X. Martinet, X. Wang, X. Wang, X. E. Tan, X. Xia, X. Xie, X. Jia, X. Wang, Y. Goldschlag, Y. Gaur, Y. Babaei, Y. Wen, Y. Song, Y. Zhang, Y. Li, Y. Mao, Z. D. Coudert, Z. Yan, Z. Chen, Z. Papakipos, A. Singh, A. Srivastava, A. Jain, A. Kelsey, A. Shajnfeld, A. Gangidi, A. Victoria, A. Goldstand, A. Menon, A. Sharma, A. Boesenberg, A. Baevski, A. Feinstein, A. Kallet, A. Sangani, A. Teo, A. Yunus, A. Lupu, A. Alvarado, A. Caples, A. Gu, A. Ho, A. Poulton, A. Ryan, A. Ramchandani, A. Dong, A. Franco, A. Goyal, A. Saraf, A. Chowdhury, A. Gabriel, A. Bharambe, A. Eisenman, A. Yazdan, B. James, B. Maurer, B. Leonhardi, B. Huang, B. Loyd, B. De Paola, B. Paranjape, B. Liu, B. Wu, B. Ni, B. Hancock, B. Wasti, B. Spence, B. Stojkovic, B. Gamido, B. Montalvo, C. Parker, C. Burton, C. Mejia, C. Liu, C. Wang, C. Kim, C. Zhou, C. Hu, C.-H. Chu, C. Cai, C. Tindal, C. Feichtenhofer, C. Gao, D. Civin, D. Beaty, D. Kreymer, D. Li, D. Adkins, D. Xu, D. Testuggine, D. David, D. Parikh, D. Liskovich, D. Foss, D. Wang, D. Le, D. Holland, E. Dowling, E. Jamil, E. Montgomery, E. Presani, E. Hahn, E. Wood, E.-T. Le, E. Brinkman, E. Arcaute, E. Dunbar, E. Smothers, F. Sun, F. Kreuk, F. Tian, F. Kokkinos, F. Ozgenel, F. Caggioni, F. Kanayet, F. Seide, G. M. Florez, G. Schwarz, G. Badeer, G. Swee, G. Halpern, G. Herman, G. Sizov, Guangyi, Zhang, G. Lakshminarayanan, H. Inan, H. Shojanazeri, H. Zou, H. Wang, H. Zha, H. Habeeb, H. Rudolph, H. Suk, H. Aspegren, H. Goldman, H. Zhan, I. Damlaj, I. Molybog, I. Tufanov, I. Leontiadis, I.-E. Veliche, I. Gat, J. Weissman, J. Geboski, J. Kohli, J. Lam, J. Asher, J.-B. Gaya, J. Marcus, J. Tang, J. Chan, J. Zhen, J. Reizenstein, J. Teboul, J. Zhong, J. Jin, J. Yang, J. Cummings, J. Carvill, J. Shepard, J. McPhie, J. Torres, J. Ginsburg, J. Wang, K. Wu, K. H. U, K. Saxena, K. Khandelwal, K. Zand, K. Matosich, K. Veeraraghavan, K. Michelena, K. Li, K. Jagadeesh, K. Huang, K. Chawla, K. Huang, L. Chen, L. Garg, L. A, L. Silva, L. Bell, L. Zhang, L. Guo, L. Yu, L. Moshkovich, L. Wehrstedt, M. Khabsa, M. Avalani, M. Bhatt, M. Mankus, M. Hasson, M. Lennie, M. Reso, M. Groshev, M. Naumov, M. Lathi, M. Keneally, M. Liu, M. L. Seltzer, M. Valko, M. Restrepo, M. Patel, M. Vyatskov, M. Samvelyan, M. Clark, M. Macey, M. Wang, M. J. Hermoso, M. Metanat, M. Rastegari, M. Bansal, N. Santhanam, N. Parks, N. White, N. Bawa, N. Singhal, N. Egebo, N. Usunier, N. Mehta, N. P. Laptev, N. Dong, N. Cheng, O. Chernoguz, O. Hart, O. Salpekar, O. Kalinli, P. Kent, P. Parekh, P. Saab, P. Balaji, P. Rittner, P. Bontrager, P. Roux, P. Dollar, P. Zvyagina, P. Ratanchandani, P. Yuvraj, Q. Liang, R. Alao, R. Rodriguez, R. Ayub, R. Murthy, R. Nayani, R. Mitra, R. Parthasarathy, R. Li, R. Hogan, R. Battey, R. Wang, R. Howes, R. Rinott, S. Mehta, S. Siby, S. J. Bondu, S. Datta, S. Chugh, S. Hunt, S. Dhillon, S. Sidorov, S. Pan, S. Mahajan, S. Verma, S. Yamamoto, S. Ramaswamy, S. Lindsay, S. Lindsay, S. Feng, S. Lin, S. C. Zha, S. Patil, S. Shankar, S. Zhang, S. Zhang, S. Wang, S. Agarwal, S. Sajuyigbe, S. Chintala, S. Max, S. Chen, S. Kehoe, S. Satterfield, S. Govindaprasad, S. Gupta, S. Deng, S. Cho, S. Virk, S. Subramanian, S. Choudhury, S. Goldman, T. Remez, T. Glaser, T. Best, T. Koehler, T. Robinson, T. Li, T. Zhang, T. Matthews, T. Chou, T. Shaked, V. Vontimitta, V. Ajayi, V. Montanez, V. Mohan, V. S. Kumar, V. Mangla, V. Ionescu, V. Poenaru, V. T. Mihailescu, V. Ivanov, W. Li, W. Wang, W. Jiang, W. Bouaziz, W. Constable, X. Tang, X. Wu, X. Wang, X. Wu, X. Gao, Y. Kleinman, Y. Chen, Y. Hu, Y. Jia, Y. Qi, Y. Li, Y. Zhang, Y. Zhang, Y. Adi, Y. Nam, Yu, Wang, Y. Zhao, Y. Hao, Y. Qian, Y. Li, Y. He, Z. Rait, Z. DeVito, Z. Rosnbrick, Z. Wen, Z. Yang, Z. Zhao, Z. Ma, The Llama 3 Herd of Models. (2024).

68. T. Fornaciari, F. Bianchi, M. Poesio, D. Hovy, BERTective: Language Models and Contextual Information for Deception Detection. *EACL 2021 - 16th Conference of the European Chapter of the Association for Computational Linguistics, Proceedings of the Conference*, 2699–2708 (2021).

69. R. Loconte, R. Russo, P. Capuozzo, P. Pietrini, G. Sartori, Verbal lie detection using Large Language Models. *Scientific Reports 2023 13:1* **13**, 1–19 (2023).

70. M. Schweinsberg, M. Feldman, N. Staub, O. R. van den Akker, R. C. M. van Aert, M. A. L. M. van Assen, Y. Liu, T. Althoff, J. Heer, A. Kale, Z. Mohamed, H. Amireh, V. Venkatesh Prasad, A. Bernstein, E. Robinson, K. Snellman, S. Amy Sommer, S. M. G. Otner, D. Robinson, N. Madan, R. Silberzahn, P. Goldstein, W. Tierney, T. Murase, B. Mandl, D. Viganola, C. Strobl, C. B. C. Schaumans, S. Kelchtermans, C. Naseeb, S. Mason Garrison, T. Yarkoni, C. S. Richard Chan, P. Adie, P. Alaburda, C. Albers, S. Alspaugh, J. Alstott, A. A. Nelson, E. Ariño de la Rubia, A. Arzi, Š. Bahník, J. Baik, L. Winther Balling, S. Banker, D. AA Baranger, D. J. Barr, B. Barros-Rivera, M. Bauer, E. Blaise, L. Boelen, K. Bohle Carbonell, R. A. Briers, O. Burkhard, M. A. Canela, L. Castrillo, T. Catlett, O. Chen, M. Clark, B. Cohn, A. Coppock, N. Cugueró-Escofet, P. G. Curran, W. Cyrus-Lai, D. Dai, G. Valentino Dalla Riva, H. Danielsson, R. de F. S. M. Russo, N. de Silva, C. Derungs, F. Dondelinger, C. Duarte de Souza, B. Tyson Dube, M. Dubova, B. Mark Dunn, P. Adriaan Edelsbrunner, S. Finley, N.

Fox, T. Gnambs, Y. Gong, E. Grand, B. Greenawalt, D. Han, P. H. P. Hanel, A. B. Hong, D. Hood, J. Hsueh, L. Huang, K. N. Hui, K. A. Hultman, A. Javaid, L. Ji Jiang, J. Jong, J. Kamdar, D. Kane, G. Kappler, E. Kaszubowski, C. M. Kavanagh, M. Khabsa, B. Kleinberg, J. Kouros, H. Krause, A. M. Krypotos, D. Lavbič, R. Ling Lee, T. Leffel, W. Yang Lim, S. Liverani, B. Loh, D. Lønsmann, J. Wei Low, A. Lu, K. MacDonald, C. R. Madan, L. Hjorth Madsen, C. Maimone, A. Mangold, A. Marshall, H. Ester Matskewich, K. Mavon, K. L. McLain, A. A. McNamara, M. McNeill, U. Mertens, D. Miller, B. Moore, A. Moore, E. Nantz, Z. Nasrullah, V. Nejkovic, C. S. Nell, A. Arthur Nelson, G. Nilsonne, R. Nolan, C. E. O'Brien, P. O'Neill, K. O'Shea, T. Olita, J. Otterbacher, D. Palsetia, B. Pereira, I. Pozdniakov, J. Protzko, J. N. Reyt, T. Riddle, A. (Akmal) Ridhwan Omar Ali, I. Ropovik, J. M. Rosenberg, S. Rothen, M. Schulte-Mecklenbeck, N. Sharma, G. Shotwell, M. Skarzynski, W. Stedden, V. Stodden, M. A. Stoffel, S. Stoltzman, S. Subbaiah, R. Tatman, P. H. Thibodeau, S. Tomkins, A. Valdivia, G. B. Druijff-van de Woestijne, L. Viana, F. Villesèche, W. Duncan Wadsworth, F. Wanders, K. Watts, J. D. Wells, C. E. Whelpley, A. Won, L. Wu, A. Yip, C. Youngflesh, J. C. Yu, A. Zandian, L. Zhang, C. Zibman, E. Luis Uhlmann, Same data, different conclusions: Radical dispersion in empirical results when independent analysts operationalize and test the same hypothesis. *Organ. Behav. Hum. Decis. Process.* **165**, 228–249 (2021).

71. B. Kleinberg, A. Arntz, B. Verschuere, Being accurate about accuracy in verbal deception detection. *PLoS One* **14**, e0220228 (2019).
72. F. Pargent, R. Schoedel, C. Stachl, Best Practices in Supervised Machine Learning: A Tutorial for Psychologists. *Adv. Methods Pract. Psychol. Sci.* **6** (2023).
73. A. Velutharambath, K. Sassenberg, R. Klinger, What if Deception Cannot be Detected? A Cross-Linguistic Study on the Limits of Deception Detection from Text. (2025).
74. C. Dwork, V. Feldman, M. Hardt, T. Pitassi, O. Reingold, A. Roth, The reusable holdout: Preserving validity in adaptive data analysis. *Science (1979).* **349**, 636–638 (2015).
75. F. Anvari, D. Lakens, The replicability crisis and public trust in psychological science. *Compr. Results Soc. Psychol.* **3**, 266–286 (2018).
76. M. Mitchell, S. Wu, A. Zaldivar, P. Barnes, L. Vasserman, B. Hutchinson, E. Spitzer, I. D. Raji, T. Gebru, Model cards for model reporting. *FAT* 2019 - Proceedings of the 2019 Conference on Fairness, Accountability, and Transparency*, 220–229 (2019).
77. J. M. Hofman, A. Chatzimparmpas, A. Sharma, D. J. Watts, J. Hullman, Pre-registration for Predictive Modeling. (2023).
78. K. L. Milkman, D. Gromet, H. Ho, J. S. Kay, T. W. Lee, P. Pandiloski, Y. Park, A. Rai, M. Bazerman, J. Beshears, L. Bonacorsi, C. Camerer, E. Chang, G. Chapman, R. Cialdini, H. Dai, L. Eskreis-Winkler, A. Fishbach, J. J. Gross, S. Horn, A. Hubbard, S. J. Jones, D. Karlan, T. Kautz, E. Kirgios, J. Klusowski, A. Kristal, R. Ladhania, G. Loewenstein, J. Ludwig, B. Mellers, S. Mullainathan, S. Saccardo, J. Spiess, G. Suri, J. H. Talloen, J. Taxer, Y. Trope, L. Ungar, K. G. Volpp, A. Whillans, J. Zinman, A. L. Duckworth, Megastudies improve the impact of applied behavioural science. *Nature 2021 600:7889* **600**, 478–483 (2021).
79. M. J. Salganik, I. Lundberg, A. T. Kindel, C. E. Ahearn, K. Al-Ghoneim, A. Almaatouq, D. M. Altschul, J. E. Brand, N. B. Carnegie, R. J. Compton, D. Datta, T. Davidson, A. Filippova, C. Gilroy, B. J. Goode, E. Jahani, R. Kashyap, A. Kirchner, S. McKay, A. C. Morgan, A. Pentland, K. Polimis, L. Raes, D. E. Rigobon, C. V. Roberts, D. M. Stanescu, Y. Suhara, A. Usmani, E. H. Wang, M. Adem, A. Alhajri, B. AlShebli, R. Amin, R. B. Amos, L. P. Argyle, L. Baer-Bositis, M. Büchi, B. R. Chung, W. Eggert, G. Faletto, Z. Fan, J. Freese, T. Gadgil, J. Gagné, Y. Gao, A. Halpern-Manners, S. P. Hashim, S. Hausen, G. He, K. Higuera, B. Hogan, I. M. Horwitz, L. M. Hummel, N. Jain, K. Jin, D. Jurgens, P. Kaminski, A. Karapetyan, E. H. Kim, B. Leizman, N. Liu, M. Möser, A. E. Mack, M. Mahajan, N. Mandell, H. Marahrens, D. Mercado-Garcia, V. Mocz, K. Mueller-Gastell, A. Musse, Q. Niu, W. Nowak, H. Omidvar, A. Or, K. Ouyang, K. M. Pinto, E. Porter, K. E. Porter, C. Qian, T. Rauf, A. Sargsyan, T. Schaffner, L. Schnabel, B. Schonfeld, B. Sender, J. D. Tang, E. Tsurkov, A. van Loon, O. Varol, J. Wang, F. Wang, S. Weissman, K. Whitaker, M. K. Wolters, W. L. Woon, X. Wang, Z. Wang, J. Wu, C. Wu, K. Yang, J. Yin, B. Zhao, C. Zhu, J. Brooks-Gunn, B. E. Engelhardt, M. Hardt, D. Knox, K. Levy, A. Narayanan, B. M. Stewart, D. J. Watts, S. McLanahan, Measuring the predictability of life outcomes with a scientific mass collaboration. *Proc. Natl. Acad. Sci. U. S. A.* **117**, 8398–8403 (2020).
80. G. Bogaard, E. H. Meijer, A. Vrij, H. Merckelbach, Strong, but Wrong: Lay People's and Police Officers' Beliefs about Verbal and Nonverbal Cues to Deception. *PLoS One* **11**, e0156615 (2016).
81. B. Verschuere, C. C. Lin, S. Huismann, B. Kleinberg, M. Willemse, E. C. J. Mei, T. van Goor, L. H. S. Löwy, O. K. Appiah, E. Meijer, The use-the-best heuristic facilitates deception detection. *Nature Human Behaviour 2023 7:5* **7**, 718–728 (2023).
82. B. Kleinberg, B. Verschuere, How humans impair automated deception detection performance. *Acta Psychol. (Amst).* **213** (2021).

83. R. Loconte, M. Monaro, P. Pietrini, B. Verschuere, B. Kleinberg, Humans incorrectly reject confident accusatory AI judgments. *Comput. Human Behav.* **182**, 109019 (2026).
84. A. von Schenk, V. Klockmann, J. F. Bonnefon, I. Rahwan, N. Köbis, Lie detection algorithms disrupt the social dynamics of accusation behavior. *iScience* **27**, 110201 (2024).
85. R. Bommasani, D. A. Hudson, E. Adeli, R. Altman, S. Arora, S. von Arx, M. S. Bernstein, J. Bohg, A. Bosselut, E. Brunskill, E. Brynjolfsson, S. Buch, D. Card, R. Castellon, N. Chatterji, A. Chen, K. Creel, J. Q. Davis, D. Demszky, C. Donahue, M. Doumbouya, E. Durmus, S. Ermon, J. Etchemendy, K. Ethayarajh, L. Fei-Fei, C. Finn, T. Gale, L. Gillespie, K. Goel, N. Goodman, S. Grossman, N. Guha, T. Hashimoto, P. Henderson, J. Hewitt, D. E. Ho, J. Hong, K. Hsu, J. Huang, T. Icard, S. Jain, D. Jurafsky, P. Kalluri, S. Karamcheti, G. Keeling, F. Khani, O. Khattab, P. W. Koh, M. Krass, R. Krishna, R. Kuditipudi, A. Kumar, F. Ladhak, M. Lee, T. Lee, J. Leskovec, I. Levent, X. L. Li, X. Li, T. Ma, A. Malik, C. D. Manning, S. Mirchandani, E. Mitchell, Z. Munyikwa, S. Nair, A. Narayan, D. Narayanan, B. Newman, A. Nie, J. C. Niebles, H. Nilforoshan, J. Nyarko, G. Ogut, L. Orr, I. Papadimitriou, J. S. Park, C. Piech, E. Portelance, C. Potts, A. Raghunathan, R. Reich, H. Ren, F. Rong, Y. Roohani, C. Ruiz, J. Ryan, C. Ré, D. Sadigh, S. Sagawa, K. Santhanam, A. Shih, K. Srinivasan, A. Tamkin, R. Taori, A. W. Thomas, F. Tramèr, R. E. Wang, W. Wang, B. Wu, J. Wu, Y. Wu, S. M. Xie, M. Yasunaga, J. You, M. Zaharia, M. Zhang, T. Zhang, X. Zhang, Y. Zhang, L. Zheng, K. Zhou, P. Liang, On the Opportunities and Risks of Foundation Models. (2021).
86. M. Binz, E. Akata, M. Bethge, F. Brändle, F. Callaway, J. Coda-Forno, P. Dayan, C. Demircan, M. K. Eckstein, N. Éltető, T. L. Griffiths, S. Haridi, A. K. Jagadish, L. Ji-An, A. Kipnis, S. Kumar, T. Ludwig, M. Mathony, M. Mattar, A. Modirshanechi, S. S. Nath, J. C. Peterson, M. Rmus, E. M. Russek, T. Saanum, J. A. Schubert, L. M. Schulze Buschoff, N. Singhi, X. Sui, M. Thalmann, F. J. Theis, V. Truong, V. Udandarao, K. Voudouris, R. Wilson, K. Witte, S. Wu, D. U. Wulff, H. Xiong, E. Schulz, A foundation model to predict and capture human cognition. *Nature 2025 644:8078* **644**, 1002–1009 (2025).
87. M. J. Page, J. E. McKenzie, P. M. Bossuyt, I. Boutron, T. C. Hoffmann, C. D. Mulrow, L. Shamseer, J. M. Tetzlaff, E. A. Akl, S. E. Brennan, R. Chou, J. Glanville, J. M. Grimshaw, A. Hróbjartsson, M. M. Lalu, T. Li, E. W. Loder, E. Mayo-Wilson, S. McDonald, L. A. McGuinness, L. A. Stewart, J. Thomas, A. C. Tricco, V. A. Welch, P. Whiting, D. Moher, The PRISMA 2020 statement: an updated guideline for reporting systematic reviews. *BMJ* **372** (2021).
88. R. van de Schoot, J. de Bruin, R. Schram, P. Zahedi, J. de Boer, F. Weijdema, B. Kramer, M. Huijts, M. Hoogerwerf, G. Ferdinands, A. Harkema, J. Willemsen, Y. Ma, Q. Fang, S. Hindriks, L. Tummers, D. L. Oberski, An open source machine learning framework for efficient and transparent systematic reviews. *Nature Machine Intelligence 2021 3:2* **3**, 125–133 (2021).
89. J. Boetje, R. van de Schoot, The SAFE procedure: a practical stopping heuristic for active learning-based screening in systematic reviews and meta-analyses. *Syst. Rev.* **13**, 1–10 (2024).
90. J. J. Teijema, L. Hofstee, M. Brouwer, J. de Bruin, G. Ferdinands, J. de Boer, P. Vizan, S. van den Brand, C. Bockting, R. van de Schoot, A. Bagheri, Active learning-based systematic reviewing using switching classification models: the case of the onset, maintenance, and relapse of depressive disorders. *Front. Res. Metr. Anal.* **8** (2023).
91. W. Viechtbauer, Conducting Meta-Analyses in R with the metafor Package. *J. Stat. Softw.* **36**, 1–48 (2010).
92. A. E. Galeotti, C. Meini, Scientific Misinformation and Fake News: A Blurred Boundary. *Soc. Epistemol.* **36**, 703–718 (2022).
93. R. van de Schoot, B. M. Coimbra, T. Evenhuis, P. Lombaers, F. Weijdema, L. de Bruin, R. Neeleman, E. Grandfield, M. Sijbrandij, J. J. Teijema, E. Jalsovec, M. P. Bron, S. Winter, J. de Bruin, M. van Zuiden, The hunt for the last relevant paper: blending the best of humans and AI. *Eur. J. Psychotraumatol.* **16**, 2546214 (2025).
94. F. Soldner, B. Kleinberg, S. D. Johnson, Confounds and overestimations in fake review detection: Experimentally controlling for product-ownership and data-origin. *PLoS One* **17**, e0277869 (2022).
95. J. Li, M. Ott, C. Cardie, E. Hovy, Towards a General Rule for Identifying Deceptive Opinion Spam. *52nd Annual Meeting of the Association for Computational Linguistics, ACL 2014 - Proceedings of the Conference* **1**, 1566–1576 (2014).

## References of reports included in this review

1. Feng, S., Banerjee, R. & Choi, Y. Syntactic Stylometry for Deception Detection. 171–175 Preprint at https://aclanthology.org/P12-2034/ (2012).
2. Ren, Y. & Zhang, Y. Deceptive Opinion Spam Detection Using Neural Network. 140–150 Preprint at https://aclanthology.org/C16-1014/ (2016).
3. Kleinberg, B., Loconte, R. & Verschuere, B. Effective faking of verbal deception detection with target-aligned adversarial attacks. *Legal Criminol. Psychol.* **31**, 249–272 (2026).
4. Catelli, R., Fujita, H., De Pietro, G. & Esposito, M. Deceptive reviews and sentiment polarity: Effective link by exploiting BERT. *Expert Syst. Appl.* **209**, (2022).
5. Chebbi, S. & Jebara, S. Ben. Deception detection using multimodal fusion approaches. *Multimed. Tools Appl.* **82**, 13073–13102 (2023).
6. Dunbar, N. E. *et al.* Detecting ulterior motives from verbal cues in group deliberations. *Front. Psychol.* **14**, (2023).
7. Wawer, A., Wojdyga, G., Sarzy´nska, J. & Sarzy´nska-Wawer, S. *Fact Checking or Psycholinguistics: How to Distinguish Fake and True Claims?* https://github.com/goldsmith/ (2019).
8. Hazra, S. & Majumder, B. P. To Tell the Truth: Language of Deception and Language Models. in *Proceedings of the 2024 Conference of the North American Chapter of the Association for Computational Linguistics* vol. 1 8506–8520 (Long Papers, 2024).
9. Fukarel, S. G., Edriatama, O. P., Aqila, M. R., Masaling, N. A. P. & Chowanda, A. Phishing Detection with BERT: A Transformer-Based Approach for Cybersecurity. in *Digest of Technical Papers - IEEE International Conference on Consumer Electronics* 265–271 (Institute of Electrical and Electronics Engineers Inc., 2025). doi:10.1109/ISCT66099.2025.11297271.
10. Messal, M., Miah, M., Anika, A., Shi, X. & Huang, R. Hidden in Plain Sight: Evaluation of the Deception Detection Capabilities of LLMs in Multimodal Settings. in *Proceedings of the 63rd Annual Meeting of the Association for Computational Linguistics* vol. 1 31013–31034 (2025).
11. Quijano-Sánchez, L., Liberatore, F., Camacho-Collados, J. & Camacho-Collados, M. Applying automatic text-based detection of deceptive language to police reports: Extracting behavioral patterns from a multi-step classification model to understand how we lie to the police. *Knowl. Based. Syst.* **149**, 155–168 (2018).
12. Dulaney, E. F. *CHANGES IN LANGUAGE BEHAVIOR AS A FUNCTION OF VERACITY*. *Human Communication Research Fall* vol. 9 https://academic.oup.com/hcr/article/9/1/75/4587896 (1982).
13. Rill-García, R., Villaseñor-Pineda, L., Reyes-Meza, V. & Escalante, H. J. From Text to Speech: A Multimodal Cross-Domain Approach for Deception Detection. in *Lecture Notes in Computer Science (including subseries Lecture Notes in Artificial Intelligence and Lecture Notes in Bioinformatics)* vol. 11188 LNCS 164–177 (Springer Verlag, 2019).
14. Jensen, M. L., Meservy, T. O., Burgoon, J. K. & Nunamaker, J. F. Automatic, multimodal evaluation of human interaction. *Group Decis. Negot.* **19**, 367–389 (2010).
15. Mbaziira, A. V. & Jones, J. H. Hybrid text-based deception models for native and non-native English cybercriminal networks. in *ACM International Conference Proceeding Series* vol. Part F130280 23–27 (Association for Computing Machinery, 2017).
16. Songram, P., Choompol, A., Thipsanthia, P. & Boonjing Veera. Detecting Thai messages leading to deception on Facebook. in *International Symposium on Integrated Uncertainty in Knowledge Modelling and Decision Making* 293–304 (Springer International Publishing, 2016).
17. Triplett, S., Minami, S. & Verma, R. Effects of Soft-Domain Transfer and Named Entity Information on Deception Detection. http://arxiv.org/abs/2410.14814 (2024).
18. Sarzynska-Wawer, J., Pawlak, A., Szymanowska, J., Hanusz, K. & Wawer, A. Truth or lie: Exploring the language of deception. *PLoS One* **18**, (2023).
19. Alawadh, H. M., Alabrah, A., Meraj, T. & Rauf, H. T. Semantic Features-Based Discourse Analysis Using Deceptive and Real Text Reviews. *Information (Switzerland)* **14**, (2023).
20. Soldner, F., Pérez-Rosas, V. & Mihalcea, R. Box of Lies: Multimodal Deception Detection in Dialogues. in *Proceedings of the 2019 conference of the North American chapter of the association for computational linguistics: human language technologies* 1768–1777 (Association for Computational Linguistics, 2019).
21. Vrljic, B. *et al.* Would I Lie to You? Using Large Language Models for Deception Detection. in *2025 7th International Conference on Natural Language Processing, ICNLP 2025* 595–600 (Institute of Electrical and Electronics Engineers Inc., 2025). doi:10.1109/ICNLP65360.2025.11108620.

22. Venkatesh, S., Ramachandra, R. & Bours, P. Robust Algorithm for Multimodal Deception Detection. in *Proceedings - 2nd International Conference on Multimedia Information Processing and Retrieval, MIPR 2019* 534–537 (Institute of Electrical and Electronics Engineers Inc., 2019). doi:10.1109/MIPR.2019.00108.
23. Abdulridha, F. & Albaker, B. M. Non-invasive real-time multimodal deception detection using machine learning and parallel computing techniques. *Soc. Netw. Anal. Min.* **14**, (2024).
24. Belbachir, F. & Alkan, A. K. Features influencing the concept of trust in online reviews. in *2022 17th Iberian Conferences on Information Systems and Technologies* (IEEE, 2022).
25. Briscoe, E. J., Appling, D. S. & Hayes, H. Cues to deception in social media communications. in *Proceedings of the Annual Hawaii International Conference on System Sciences* 1435–1443 (IEEE Computer Society, 2014). doi:10.1109/HICSS.2014.186.
26. Boumber, D., Qachfar, F. Z. & Verma, R. M. *Domain-Agnostic Adapter Architecture for Deception Detection: Extensive Evaluations with the DIFrauD Benchmark*. https://huggingface.co/datasets/.
27. Abdelhamid, S. E., Davis, B. & Le, D. K. GPT-4 Meets TF-IDF: A Hybrid Approach for Detecting Spam Emails Using Machine Learning. in *2025 IEEE 15th Annual Computing and Communication Workshop and Conference, CCWC 2025* 619–625 (Institute of Electrical and Electronics Engineers Inc., 2025). doi:10.1109/CCWC62904.2025.10903847.
28. Al-Adhaileh, M. H. & Alsaade, F. W. Detecting and analysing fake opinions using artificial intelligence algorithms. *Intelligent Automation and Soft Computing* **32**, 643–655 (2022).
29. Chang, T., Hsu, P. Y., Cheng, M. S., Chung, C. Y. & Chung, Y. L. Detecting fake review with rumor model - case study in hotel review. in *Lecture Notes in Computer Science (including subseries Lecture Notes in Artificial Intelligence and Lecture Notes in Bioinformatics)* vol. 9243 181–192 (Springer Verlag, 2015).
30. Yadav, A., Yadav, R. & Kiruthika, M. SMS Spam Detection with NLP and Ensemble Learning for Enhanced Classification. in *2025 International Conference on Computing Technologies and Data Communication, ICCTDC 2025* (Institute of Electrical and Electronics Engineers Inc., 2025). doi:10.1109/ICCTDC64446.2025.11159039.
31. Fontanarava, J., Pasi, G. & Viviani, M. An ensemble method for the credibility assessment of user-generated content. in *Proceedings - 2017 IEEE/WIC/ACM International Conference on Web Intelligence, WI 2017* 863–868 (Association for Computing Machinery, Inc, 2017). doi:10.1145/3106426.3106464.
32. Nguyen, T., Abri, F., Siami Namin, A. & Jones, K. S. Multimodal Deception Detection Using Linguistic and Acoustic Features. in *Machine Learning, Deep Learning and AI for Cybersecurity* 565–598 (Springer Science+Business Media, 2025). doi:10.1007/978-3-031-83157-7_20.
33. Kim, S., Lee, S., Park, D. & Kang, J. Constructing and evaluating a novel crowdsourcing-based paraphrased opinion spam dataset. in *26th International World Wide Web Conference, WWW 2017* 827–836 (International World Wide Web Conferences Steering Committee, 2017). doi:10.1145/3038912.3052607.
34. Abouelenien, M., Perez-Rosas, V., Mihalcea, R. & Burzo, M. Detecting Deceptive Behavior via Integration of Discriminative Features from Multiple Modalities. *IEEE Transactions on Information Forensics and Security* **12**, 1042–1055 (2017).
35. Abdulqader, M., Namoun, A. & Alsaawy, Y. Fake Online Reviews: A unified detection model using deception theories. *IEEE Access* **10**, 128622–128655 (2022).
36. Kleinberg, B. & Verschuere, B. How humans impair automated deception detection performance. *Acta Psychol. (Amst).* **213**, (2021).
37. Kennedy, S., Walsh, N., Sloka, K., Mccarren, A. & Foster, J. *Fact or Factitious? Contextualized Opinion Spam Detection*. https://www.yelp.com/factsheet (2019).
38. Almela, Á. A corpus-based study of linguistic deception in Spanish. *Applied Sciences (Switzerland)* **11**, (2021).
39. Martinez-Torres, M. R. & Toral, S. L. A machine learning approach for the identification of the deceptive reviews in the hospitality sector using unique attributes and sentiment orientation. *Tour. Manag.* **75**, 393–403 (2019).
40. Lohumi, P., Gupta, K., Sharma, V. & Jain, R. Enhancing the Detection of Deceptive Job Advertisements: A Machine Learning Approach with Advanced Oversampling and Vectorization Techniques. in *Lecture Notes in Networks and Systems* vol. 1270 LNNS 105–122 (Springer Science and Business Media Deutschland GmbH, 2025).
41. Hanks, C. & Verma, R. M. Data Quality and Linguistic Cues for Domain-independent Deception Detection. in *Proceedings - 2022 IEEE/ACM 9th International Conference on Big Data Computing, Applications and Technologies, BDCAT 2022* 248–258 (Institute of Electrical and Electronics Engineers Inc., 2022). doi:10.1109/BDCAT56447.2022.00042.
42. Caruccio, L. *et al.* Identifying fake reviews for refund purposes: Evaluating the effectiveness of a transfer-learning model against emerging Large Language Models. *Eng. Appl. Artif. Intell.* **162**, (2025).
43. Barsever, D., Singh, S. & Neftci Emre. Building a Better Lie Detector with BERT: The Difference Between Truth and Lies. in *2020 International Joint Conference on Neural Networks* 1–7 (IEEE, 2020).

44. Ho, S. M., Hancock, J. T., Booth, C. & Liu, X. Computer-Mediated Deception: Strategies Revealed by Language-Action Cues in Spontaneous Communication. *Journal of Management Information Systems* **33**, 393–420 (2016).
45. Bhangale, S. & Roy, P. K. Is It genuine or fake? Analyzing e-commerce reviews using large language models. *Knowl. Based. Syst.* **330**, (2025).
46. Altan, I., Bachir, A., Parbhulkar, Y., Rizvi, A. M. & Farazi, M. Dual-Path Phishing Detection: Integrating Transformer-Based NLP with Structural URL Analysis. http://arxiv.org/abs/2509.20972 (2025).
47. Ennaouri, M. & Zellou, A. A scoring approach for detecting fake reviews using MRCS similarity metric enhanced by personalized k-means. *Bulletin of Electrical Engineering and Informatics* **14**, 587–595 (2025).
48. Samineni, L. *et al.* Evaluation of AI Techniques for Detecting Deceptive Reviews in Cyberspace: A Study of Pre- and Post-COVID-19 Trends. in *2023 Second International Conference on Electronics and Renewable Systems (ICEARS)* 961–967 (IEEE, 2023).
49. Li, S. & Cheng, W. Augmenting the global semantic information between words to heterogeneous graph for deception detection. *Neural Comput. Appl.* **34**, 19079–19090 (2022).
50. Fuller, C. M., Biros, D. P. & Delen, D. An investigation of data and text mining methods for real world deception detection. *Expert Syst. Appl.* **38**, 8392–8398 (2011).
51. Etaiwi, W. & Naymat, G. The Impact of applying Different Preprocessing Steps on Review Spam Detection. in *Procedia Computer Science* vol. 113 273–279 (Elsevier B.V., 2017).
52. Baldivas, R. I. A. *et al.* LegalEye: Multimodal Court Deception Detection Across Multiple Languages. *Behavioral Sciences* **15**, (2025).
53. Sharma, N. *et al.* Gradient Boosting Decision Trees for Real-Time Phishing Attack Prevention in Cybersecurity. in *Proceedings - IEEE 10th International Conference on Smart Structures and Systems, ICSSS 2025* (Institute of Electrical and Electronics Engineers Inc., 2025). doi:10.1109/ICSSS66939.2025.11346380.
54. Zhang, Z. *et al.* Metapath and syntax-aware heterogeneous subgraph neural networks for spam review detection. *Appl. Soft Comput.* **128**, (2022).
55. Himdi, H. & Alhayan, F. Optimized ensemble stacking approaches to detect Arabic phishing email. *Journal of Engineering Research (Kuwait)* **14**, 776–788 (2026).
56. Mohawesh, R., AlQarni, A. A., Alkhushayni, S. M., Daradkeh, T. & Bany Salameh, H. A new multilingual framework for fake reviews detection based on a large language model. *Journal of Supercomputing* **81**, (2025).
57. Kuwar, B. V., Maurya, B. B. P., Gupta, P. & Choudhury, N. PU-Lie: Lightweight Deception Detection in Imbalanced Diplomatic Dialogues via Positive-Unlabeled Learning. *arXiv preprint* arxiv:2507.09157 (2025).
58. Warnita, T. & Lestari, D. P. Identifying Deception in Indonesian Transcribed Interviews through Lexical-based Approach. in *Proceedings of the 31st Pacific Asia Conference on Language, Information and Computation* 148–154 (2017).
59. Tanvir, M. *et al.* Bengali Fake Review Detection using Semi-supervised Generative Adversarial Networks. in *2023 5th International Conference on Natural Language Processing (ICNLP)* 12–16 (IEEE, 2023).
60. Hernández Fusilier, D., Montes-y-Gómez, M., Rosso, P. & Guzmán Cabrera, R. Detecting positive and negative deceptive opinions using PU-learning. *Inf. Process. Manag.* **51**, 433–443 (2015).
61. Mukherjee, A., Venkataraman, V., Liu, B. & Glance, N. *What Yelp Fake Review Filter Might Be Doing?* www.aaai.org (2013).
62. Moon, S., Kim, M. Y. & Iacobucci, D. Content analysis of fake consumer reviews by survey-based text categorization. *International Journal of Research in Marketing* **38**, 343–364 (2021).
63. Bao, Y. *et al.* Deception Detection Towards Multi-turn Question Answering with Context Selector Network. in *Lecture Notes in Computer Science (including subseries Lecture Notes in Artificial Intelligence and Lecture Notes in Bioinformatics)* vol. 13629 LNCS 304–315 (Springer Science and Business Media Deutschland GmbH, 2022).
64. Deshai, N. & Bhaskara Rao, B. Unmasking deception: a CNN and adaptive PSO approach to detecting fake online reviews. *Soft comput.* **27**, 11357–11378 (2023).
65. Molla, A., Biadgie, Y. & Sohn, K. A. Detecting negative deceptive opinion from tweets. in *Lecture Notes in Electrical Engineering* vol. 425 329–339 (Springer Verlag, 2018).
66. Toma, C. L. & Hancock, J. T. What Lies Beneath: The Linguistic Traces of Deception in Online Dating Profiles. *Journal of Communication* **62**, 78–97 (2012).
67. Veliyath, A. J., Ahmad, A., Abraham, A. B., Kancharla, S. H. H. & Chavan, C. P. Fake Job Detection using Statistical and NLP Based Analysis. in *ICEIEC 2025 - Proceedings of 2025 IEEE 15th International Conference on Electronics Information and Emergency Communication* 167–172 (Institute of Electrical and Electronics Engineers Inc., 2025). doi:10.1109/ICEIEC65904.2025.11273154.

68. Elangovan, R., Abirami, A. M. & Thirumalai, K. G. FSSHAP: Global Interpretable Feature Selection Using XAI for the Classification of Spam SMS in Dravidian Languages. *IEEE Multimedia* **32**, 92–103 (2025).
69. Fahmy, O. S., Elhaddad, K. A., Badran, K. M. & Elhadad, M. K. Enhancing Textual Deception Detection: A Fused Handcrafted Feature Approach with Machine Learning Models. in *Communications in Computer and Information Science* vol. 2487 CCIS 216–230 (Springer Science and Business Media Deutschland GmbH, 2025).
70. Songram, P. *et al.* Deep learning for classifying thai deceptive messages. *Indonesian Journal of Electrical Engineering and Computer Science* **30**, 1232–1241 (2023).
71. Hayat, U., Saeed, A., Vardag, M. H. K., Ullah, M. F. & Iqbal, N. Roman Urdu Fake Reviews Detection Using Stacked LSTM Architecture. *SN Comput. Sci.* **3**, (2022).
72. Alsaad, M. M. B. & Joshi, H. A Framework for Online Fake Review Detection on Yelp Electronics Product Dataset Using Machine Learning Techniques. in *Lecture Notes in Networks and Systems* vol. 1305 LNNS 409–423 (Springer Science and Business Media Deutschland GmbH, 2025).
73. Xu, S., Cuan, H., Yin, Z. & Yin, C. A Hybridized Approach for Enhanced Fake Review Detection. *IEEE Trans. Comput. Soc. Syst.* **11**, 7448–7466 (2024).
74. Chou, H. C., Liu, Y. W. & Lee, C. C. Automatic Deception Detection using Multiple Speech and Language Communicative Descriptors in Dialogs. *APSIPA Trans. Signal Inf. Process.* **10**, (2021).
75. Alwayle, I. M. *et al.* Automated Spam Review Detection Using Hybrid Deep Learning on Arabic Opinions. *Computer Systems Science and Engineering* **46**, 2947–2961 (2023).
76. Wu, Z., Singh, B., Davis, L. S. & Subrahmanian, V. S. Deception Detection in Videos. http://arxiv.org/abs/1712.04415 (2017).
77. Singh, A. & Chatterjee, K. A Comparative Approach for Opinion Spam Detection Using Sentiment Analysis. in *Lecture Notes in Networks and Systems* vol. 329 511–522 (Springer Science and Business Media Deutschland GmbH, 2022).
78. Pérez-Rosas, V., Abouelenien, M., Mihalcea, R. & Burzo, M. Deception detection using real-life trial data. in *ICMI 2015 - Proceedings of the 2015 ACM International Conference on Multimodal Interaction* 59–66 (Association for Computing Machinery, Inc, 2015). doi:10.1145/2818346.2820758.
79. Lozano, M. G. & Fernquist, J. Identifying Deceptive Reviews: Feature Exploration, Model Transferability and Classification Attack. in *European Intelligence and Security Informatics Conference (EISIC)* 109–116 (IEEE, 2019).
80. Zaki, N. *et al.* Node embedding approach for accurate detection of fake reviews: a graph-based machine learning approach with explainable AI. *Int. J. Data Sci. Anal.* **18**, 295–315 (2024).
81. Asaad, W. H., Allami, R. & Ali, Y. H. Fake Review Detection Using Machine Learning. *Revue d'Intelligence Artificielle* **37**, 1159–1166 (2023).
82. Zhong, M. *et al.* Fast Detection of Deceptive Reviews by Combining the Time Series and Machine Learning. *Complexity* **2021**, (2021).
83. Bicer, B. & Dibeklioglu, H. Automatic Deceit Detection Through Multimodal Analysis of High-Stake Court-Trials. *IEEE Trans. Affect. Comput.* **15**, 342–356 (2024).
84. Cao, N., Ji, S., Chiu, D. K. W., He, M. & Sun, X. A deceptive review detection framework: Combination of coarse and fine-grained features. *Expert Syst. Appl.* **156**, (2020).
85. Catelli, R. *et al.* A new Italian Cultural Heritage data set: detecting fake reviews with BERT and ELECTRA leveraging the sentiment. *IEEE Access* **11**, 52214–52225 (2023).
86. Yoo, B. & Kim, K. J. Finding deceivers in social context with large language models and how to find them: the case of the Mafia game. *Sci. Rep.* **14**, (2024).
87. Zhang, J., Levitan, S. I. & Hirschberg, J. Multimodal deception detection using automatically extracted acoustic, visual, and lexical features. in *Proceedings of the Annual Conference of the International Speech Communication Association, INTERSPEECH* vols 2020-October 359–363 (International Speech Communication Association, 2020).
88. Zhou, L., Shi, Y. & Zhang, D. A statistical language modeling approach to online deception detection. in *IEEE Transactions on Knowledge and Data Engineering* vol. 20 1077–1081 (2008).
89. Andresini, G. *et al.* EUPHORIA: A neural multi-view approach to combine content and behavioral features in review spam detection. *Journal of Computational Mathematics and Data Science* **3**, (2022).
90. Bajaj, N. *et al.* Deception detection in conversations using the proximity of linguistic markers. *Knowl. Based. Syst.* **267**, (2023).
91. Beshay, Y., Ibrahim, N., Abu-Elkheir, M., Abdennadher, S. & Elmougy, A. Fake Emotion Detection Using Natural Language Processing. in *Communications in Computer and Information Science* vol. 2149 CCIS 93–103 (Springer Science and Business Media Deutschland GmbH, 2025).
92. Heredia, B., Khoshgoftaar, T. M., Prusa, J. D. & Crawford, M. Improving detection of untrustworthy online reviews using ensemble learners combined with feature selection. *Soc. Netw. Anal. Min.* **7**, (2017).

93. Batra, H. & Nelson, L. ESD: E-mail Spam Detection using Cybersecurity-Driven Header Analysis and Machine Learning based Content Analysis. *International Journal of Performability Engineering* **20**, 205–213 (2024).
94. Toma, C. L. & Hancock, J. T. Reading between the lines: Linguistic cues to deception in online dating profiles. in *Proceedings of the 2010 ACM conference on Computer supported cooperative work* 5–8 (2010).
95. Barushka, A. & Hajek, P. The Effect of Text Preprocessing Strategies on Detecting Fake Consumer Reviews. in *ACM International Conference Proceeding Series* 13–17 (Association for Computing Machinery, 2019). doi:10.1145/3383902.3383908.
96. Kanmani, S. & Balasubramanian, S. Leveraging Readability and Sentiment in Spam Review Filtering Using Transformer Models. *Computer Systems Science and Engineering* **45**, 1439–1454 (2023).
97. Wang, J. *et al.* Fake review detection based on multiple feature fusion and rolling collaborative training. *IEEE Access* **8**, 182625–182639 (2020).
98. Archchitha, K. & Charles, E. Y. A. Opinion Spam Detection in Online Reviews Using Neural Networks. in *2019 19th International conference on advances in ICT for emerging regions (ICTer)* 1–6 (IEEE, 2019).
99. Wawer, A. & Sarzyńska-Wawer, J. Detecting Deceptive Utterances Using Deep Pre-Trained Neural Networks. *Applied Sciences (Switzerland)* **12**, (2022).
100. Hsiao, S. W. & Sun, C. Y. LoRA-like Calibration for Multimodal Deception Detection using ATSFace Data. in *Proceedings - 2023 IEEE International Conference on Big Data, BigData 2023* 2163–2172 (Institute of Electrical and Electronics Engineers Inc., 2023). doi:10.1109/BigData59044.2023.10386630.
101. Boumber, D., Tuck, B. E., Verma, R. M. & Qachfar, F. Z. LLMs for Explainable Few-shot Deception Detection. in *IWSPA 2024 - Proceedings of the 10th ACM International Workshop on Security and Privacy Analytics* 37–47 (Association for Computing Machinery, Inc, 2024). doi:10.1145/3643651.3659898.
102. Tashtoush, Y., Al-Nemrawi, M., Masa'deh, L., Al-Qudah, B. & Alqahtani, S. Leveraging Large Language Models for Detection and Intelligent Analysis of Phishing Threats. in *2025 International Conference on Cybersecurity and AI-Based Systems, Cyber-AI 2025* 326–331 (Institute of Electrical and Electronics Engineers Inc., 2025). doi:10.1109/Cyber-AI66431.2025.11233620.
103. Bhavana, B. S., Srivastava, S., Ghosh, D., Kumar, V. & Gupta, P. Context-Aware Phishing Email Detection Using Hybrid Machine Learning and Explainable AI. in *2025 17th International Conference on Contemporary Computing, IC3 2025* (Institute of Electrical and Electronics Engineers Inc., 2025). doi:10.1109/IC366947.2025.11290431.
104. Li, F. *et al.* Deceptive Answer Prediction with User Preference Graph. in *Proceedings of the 51st Annual Meeting of the Association for Computational Linguistics* 1723–1732 (2013).
105. Crudu, M. & Lupea, M. Detecting Textual Deception in Courtroom Testimonies for Romanian Language Using Neural Network Methods. in *Proceedings - 2024 IEEE 20th International Conference on Intelligent Computer Communication and Processing Conference, ICCP 2024* (Institute of Electrical and Electronics Engineers Inc., 2024). doi:10.1109/ICCP63557.2024.10793021.
106. Zhang, J. *et al.* Research on Multifeature Fusion False Review Detection Based on DynDistilBERT-BiLSTM-CNN. *IEEE Internet Things J.* **11**, 30040–30053 (2024).
107. Chen, Y. C., Chen, L. J., Hsu, P. Y., Cheng, M. S. & Chen, Y. C. Identify Deceptive Reviews in Cross-Domain Content with BERT. in *Lecture Notes in Computer Science (including subseries Lecture Notes in Artificial Intelligence and Lecture Notes in Bioinformatics)* vol. 14893 LNCS 31–46 (Springer Science and Business Media Deutschland GmbH, 2025).
108. Zhang, W., Du, Y., Yoshida, T. & Wang, Q. DRI-RCNN: An approach to deceptive review identification using recurrent convolutional neural network. *Inf. Process. Manag.* **54**, 576–592 (2018).
109. Deeb, H. *et al.* Honesty repeats itself: comparing manual and automated coding on the veracity cues total details and redundancy. *Appl. Psycholinguist.* https://doi.org/10.1017/S0142716424000298 (2024) doi:10.1017/S0142716424000298.
110. Chiraratanasopha, B. & Chay-Intr, T. Detecting Fraud Job Recruitment Using Features Reflecting from Real-world Knowledge of Fraud. *Curr. Appl. Sci. Technol.* **22**, (2022).
111. Ibraheem, S., Zhou, G. & Denero, J. Putting the Con in Context: Identifying Deceptive Actors in the Game of Mafia. in *Proceedings of the 2022 Conference of the North America Chapter of the Association for Computational Linguistics: Human Language Technologies* 158–168 (Association for Computational Linguiistics, 2022).
112. Bahaa, M., Hany, M. & Zakaria, E. E. Advancing Automated Deception Detection: A Multimodal Approach to Feature Extraction and Analysis. http://arxiv.org/abs/2407.06005 (2024).
113. Wei, C. S., Hsu, P. Y., Huang, C. W., Cheng, M. S. & Prassida, G. F. Devising a Cross-Domain Model to Detect Fake Review Comments. in *Communications in Computer and Information Science* vol. 1287 714–725 (Springer Science and Business Media Deutschland GmbH, 2020).

114. Sen, M. U. *et al.* Multimodal Deception Detection Using Real-Life Trial Data. *IEEE Trans. Affect. Comput.* **13**, 306–319 (2022).
115. Gupta, D. *et al.* Deep Learning-Based Truthful and Deceptive Hotel Reviews. *Sustainability (Switzerland)* **16**, (2024).
116. Larcker, D. F. & Zakolyukina, A. A. Detecting Deceptive Discussions in Conference Calls. *Journal of Accounting Research* **50**, 495–540 (2012).
117. Mukherjee, S., Dutta, S. & Weikum, G. Credible Review Detection with Limited Information Using Consistency Features. in *Machine Learning and Knowledge Discovery in Databases* (eds. Frasconi, P., Landwehr, N., Manco, G. & Vreeken, J.) vol. 9852 195–213 (Springer International Publishing, Cham, 2016).
118. Baishya, D., Deka, J. J., Dey, G. & Singh, P. K. SAFER: Sentiment Analysis-Based FakE Review Detection in E-Commerce Using Deep Learning. *SN Comput. Sci.* **2**, (2021).
119. Sánchez-Junquera, J., Villaseñor-Pineda, L., Montes-y-Gómez, M., Rosso, P. & Stamatatos, E. Masking domain-specific information for cross-domain deception detection. *Pattern Recognit. Lett.* **135**, 122–130 (2020).
120. Ng, K. C., Ke, P. F., So, M. K. P. & Tam, K. Y. Augmenting fake content detection in online platforms: A domain adaptive transfer learning via adversarial training approach. *Prod. Oper. Manag.* **32**, 2101–2122 (2023).
121. Abouelenien, M. *et al.* Gender Differences in Multimodal Contact-Free Deception Detection. *IEEE Multimedia* **26**, 19–30 (2019).
122. Sharifpour, K. & Lahmiri, S. Fake Review Detection Using Rating-Sentiment Inconsistency. in *Proceedings - 22nd IEEE International Conference on Machine Learning and Applications, ICMLA 2023* 926–931 (Institute of Electrical and Electronics Engineers Inc., 2023). doi:10.1109/ICMLA58977.2023.00137.
123. Nagare, S. M. *et al.* Support Vector Machine-Based SMS Spam Detection for Mobile Devices. in *2025 International Conference on Applications of Machine Intelligence and Data Analytics, ICAMIDA 2025* (Institute of Electrical and Electronics Engineers Inc., 2025). doi:10.1109/ICAMIDA64673.2025.11209256.
124. Jaiswal, M., Tabibu, S. & Bajpai, R. The Truth and Nothing but the Truth: Multimodal Analysis for Deception Detection. in *2016 IEEE 16th International Conference on Data Mining Workshops (ICDMW)* 938–943 (IEEE, 2016).
125. Brzic, B., Boticki, I. & Bagic Babac, M. Detecting Deception Using Natural Language Processing and Machine Learning in Datasets on COVID-19 and Climate Change. *Algorithms* **16**, (2023).
126. Wang, X., Liu, K. & Zhao, J. Detecting Deceptive Review Spam via Attention-Based Neural Networks. in *National ccf conference on natural language processing and chinese computing* 866–876 (Springer International Publishing, 2017).
127. Cai, C. *et al.* MDPE: A Multimodal Deception Dataset with Personality and Emotional Characteristics. in *Proceedings of the 33rd ACM International Conference on Multimedia* vol. 2657 12957–12964 (CEUR-WS, 2025).
128. Plotkina, D., Munzel, A. & Pallud, J. Illusions of truth—Experimental insights into human and algorithmic detections of fake online reviews. *J. Bus. Res.* **109**, 511–523 (2020).
129. Spyridis, Y., Younes, J.-P., Deeb, H. & Argyriou, V. Empowering Prior to Court Legal Analysis: A Transparent and Accessible Dataset for Defensive Statement Classification and Interpretation. in *2024 35th Conference of Open Innovation Association (FRUCT)* 737–743 (IEEE, 2024).
130. Nam, D., Yasmin, J. & Zulkernine, F. Effects of Pre-trained Word Embeddings on Text-based Deception Detection. in *020 IEEE Intl Conf on Dependable, Autonomic and Secure Computing, Intl Conf on Pervasive Intelligence and Computing, Intl Conf on Cloud and Big Data Computing, Intl Conf on Cyber Science and Technology Congress (DASC/PiCom/CBDCom/CyberSciTech)* 437–443 (IEEE, 2020).
131. Ennaouri, M. & Zellou, A. Enhancing Fake Review Detection Using Linguistic Exaggeration, BERT Embeddings, and Fuzzy Logic. *IEEE Access* **13**, 135957–135968 (2025).
132. Gogate, M., Adeel, A. & Hussain, A. Deep Learning Driven Multimodal Fusion for Automated Deception Detection. in *2017 IEEE symposium series on computational intelligence (SSCI)* 1–6 (IEEE, 2017).
133. Zhao, S., Xu, Z., Liu, L., Guo, M. & Yun, J. Towards Accurate Deceptive Opinions Detection Based on Word Order-Preserving CNN. *Math. Probl. Eng.* **2018**, (2018).
134. Markey, P. M. *et al.* Artificial intelligence as a tool for detecting deception in 911 homicide calls. *J. Crim. Justice* **96**, (2025).
135. Alawadh, H. M., Alabrah, A., Meraj, T. & Rauf, H. T. Discourse analysis based credibility checks to online reviews using deep learning based discourse markers. *Comput. Speech Lang.* **78**, (2023).
136. Tomašev, N. Extracting the patterns of truthfulness from political information systems in Serbia. *Information Systems Frontiers* **19**, 109–127 (2017).
137. Grabowski, D., Luczaj, K. & Saeed, K. Multimodal Behavioral Sensors for Lie Detection: Integrating Visual, Auditory, and Generative Reasoning Cues. *Sensors* **25**, 6086 (2025).

138. Salminen, J., Mustak, M., Jung, S. G., Makkonen, H. & Jansen, B. J. Decoding deception in the online marketplace: enhancing fake review detection with psycholinguistics and transformer models. *Journal of Marketing Analytics* **14**, 731–748 (2026).
139. Loconte, R. & Kleinberg, B. Examining Embedded Lies through Computational Text Analysis. *Sci. Rep.* **15**, 26482 (2025).
140. Cao, N., Ji, S., Chiu, D. K. W. & Gong, M. A deceptive reviews detection model: Separated training of multi-feature learning and classification. *Expert Syst. Appl.* **187**, (2022).
141. Ni Min, S., Fazlida, N. & Sani, M. MESSAGE CONVERSATION BASED SOCIAL ENGINEERING ATTACK DETECTION USING MACHINE LEARNING. *J. Theor. Appl. Inf. Technol.* **31**, (2025).
142. Soldner, F., Kleinberg, B. & Johnson, S. D. Confounds and overestimations in fake review detection: Experimentally controlling for product-ownership and data-origin. *PLoS One* **17**, (2022).
143. Afroz, S., Brennan, M. & Greenstadt, R. Detecting hoaxes, frauds, and deception in writing style online. in *Proceedings - IEEE Symposium on Security and Privacy* 461–475 (Institute of Electrical and Electronics Engineers Inc., 2012). doi:10.1109/SP.2012.34.
144. Pérez-Rosas, V. *et al. Verbal and Nonverbal Clues for Real-Life Deception Detection*. http://www. (2015).
145. Touma, L., Al Horani, M., Tailouni, M., Dahabiah, A. & Al Jallad, K. Voting-based Multimodal Automatic Deception Detection. *arXiv preprint arXiv* **2307.07516**, (2023).
146. Li, L., Qin, B., Ren, W. & Liu, T. Document representation and feature combination for deceptive spam review detection. *Neurocomputing* **254**, 33–41 (2017).
147. Chen, X., Wang, X., Spitzley, L. & Nunamaker, J. Trust and deception with high stakes: Evidence from the friend or foe dataset. *Decis. Support Syst.* **173**, (2023).
148. Alsubari, S. N. *et al.* Data analytics for the identification of fake reviews using supervised learning. *Computers, Materials and Continua* **70**, 3189–3204 (2022).
149. Loconte, R. *et al.* Detecting Deception Through Linguistic Cues: From Reality Monitoring to Natural Language Processing. *J. Lang. Soc. Psychol.* **44**, 523–552 (2025).
150. Ho, S. M., Liu, X., Booth, C. & Hariharan, A. Saint or sinner? Language-action cues for modeling deception using support vector machines. in *Lecture Notes in Computer Science (including subseries Lecture Notes in Artificial Intelligence and Lecture Notes in Bioinformatics)* vol. 9708 LNCS 325–334 (Springer Verlag, 2016).
151. Prome, S. A., Rafiqul Islam, M., Asirvatham, D., Hossain Sakib, M. K. & Ari Ragavan, N. LieVis: A Visual Interactive Dashboard for Lie Detection Using Machine Learning and Deep Learning Techniques. in *2023 26th International Conference on Computer and Information Technology, ICCIT 2023* (Institute of Electrical and Electronics Engineers Inc., 2023). doi:10.1109/ICCIT60459.2023.10441173.
152. Jakupov, A., Zeddini, B., Longhi, J. & Mercadal, J. Combining Stylometric and Sentiment Mining Approaches for Deceptive Opinion Spam Detection. in *Proceedings of IEEE/ACS International Conference on Computer Systems and Applications, AICCSA* (IEEE Computer Society, 2023). doi:10.1109/AICCSA59173.2023.10479298.
153. Mohawesh, R., Tran, S., Ollington, R. & Xu, S. Analysis of concept drift in fake reviews detection. *Expert Syst. Appl.* **169**, (2021).
154. Neisari, A., Rueda, L. & Saad, S. Spam review detection using self-organizing maps and convolutional neural networks. *Comput. Secur.* **106**, (2021).
155. Pak, J. & Zhou Lina. A Comparison of Features for Automatic Deception Detection in Synchronous Computer-Mediated Communication. in *2015 IEEE International Conference on Intelligence and Security Informatics (ISI)* 141–143 (IEEE, 2015).
156. Levitan, S. I., Maredia, A. & Hirschberg, J. Linguistic Cues to Deception and Perceived Deception in Interview Dialogues. in *Proceedings of the 2018 Conference of the North American Chapter of the Association for Computational Linguistics: Human Language Technologies* 1941–1950 (2018).
157. Rubikowski, M. & Wawer, A. *LNCS 7912 - The Scent of Deception: Recognizing Fake Perfume Reviews in Polish*. *LNCS* vol. 7912 (2013).
158. Mathur, L. & Matarić, M. J. Introducing Representations of Facial Affect in Automated Multimodal Deception Detection. in *ICMI 2020 - Proceedings of the 2020 International Conference on Multimodal Interaction* 305–314 (Association for Computing Machinery, Inc, 2020). doi:10.1145/3382507.3418864.
159. Kleinberg, B., van der Toolen, Y., Vrij, A., Arntz, A. & Verschuere, B. Automated verbal credibility assessment of intentions: The model statement technique and predictive modeling. *Appl. Cogn. Psychol.* **32**, 354–366 (2018).
160. Peng, Y. C., Cheng, M. S. & Hsu, P. Y. Constructing an Integrated Classifiers for Identifying Authenticity and Sentiment Analysis- A Case of Hotel Reviews. in *Proceedings - 2020 International Computer Symposium, ICS 2020* 319–324 (Institute of Electrical and Electronics Engineers Inc., 2020). doi:10.1109/ICS51289.2020.00070.

161. Pérez-Rosas, V. & Mihalcea, R. *Experiments in Open Domain Deception Detection*. (2015).
162. Humpherys, S. L. Discriminating Fraudulent Financial Statements by Indentifying Linguistic Hedging. in *Proceedings of the Fifteenth Americas Conference on Information Systems* (AMCIS, 2009).
163. Hernández-Castañeda, Á., Calvo, H. & Gambino, O. J. Impact of polarity in deception detection. *Journal of Intelligent and Fuzzy Systems* **35**, 549–558 (2018).
164. Chandaeng, T., Kosolsombat, S. & Ratanavilisagul, C. Phishing Email Detection Using Large Language Models: A Comparative Analysis of Standalone and Retrieval Augmented Generation Models. in *Proceedings - 2025 10th International Conference on Computational Intelligence and Applications, ICCIA 2025* 133–137 (Institute of Electrical and Electronics Engineers Inc., 2025). doi:10.1109/ICCIA67866.2025.00032.
165. Andresini, G. *et al. Review Spam Detection Using Multi-View Deep Learning Combining Content and Behavioral Features*. http://ceur-ws.org (2022).
166. Rout, J. K., Singh, S., Jena, S. K. & Bakshi, S. Deceptive review detection using labeled and unlabeled data. *Multimed. Tools Appl.* **76**, 3187–3211 (2017).
167. Kleinberg, B., Mozes, M., Arntz, A. & Verschuere, B. Using Named Entities for Computer-Automated Verbal Deception Detection. *J. Forensic Sci.* **63**, 714–723 (2018).
168. Lee, K. D., Han, K. & Myaeng, S. H. Capturing word choice patterns with LDA for fake review detection in sentiment analysis. in *ACM International Conference Proceeding Series* vols 13-15-June-2016 (Association for Computing Machinery, 2016).
169. Qandos, N. *et al.* Multiscale cascaded domain-based approach for Arabic fake reviews detection in e-commerce platforms. *Journal of King Saud University - Computer and Information Sciences* **36**, (2024).
170. Harris, C. G. *Detecting Deceptive Opinion Spam Using Human Computation*. www.aaai.org (2012).
171. Siering, M., Koch, J. A. & Deokar, A. V. Detecting Fraudulent Behavior on Crowdfunding Platforms: The Role of Linguistic and Content-Based Cues in Static and Dynamic Contexts. *Journal of Management Information Systems* **33**, 421–455 (2016).
172. Bauer, T. *et al.* Use of Compression Analytics to Detect Deception. in *Proceedings of the 2023 IEEE International Conference on Behavioural and Social Computing, BESC 2023* (Institute of Electrical and Electronics Engineers Inc., 2023). doi:10.1109/BESC59560.2023.10386884.
173. Hu, S. Detecting domain-specific credibility and expertise in text and speech. in *Proceedings of the Annual Conference of the International Speech Communication Association, INTERSPEECH* vols 2020-October 4208–4212 (International Speech Communication Association, 2020).
174. Weng, C. H., Lin, K. C. & Ying, J. C. Detection of Chinese Deceptive Reviews Based on Pre-Trained Language Model. *Applied Sciences (Switzerland)* **12**, (2022).
175. Litvinova, O., Litvinova, T., Seredin, P. & Lyell, J. *Deception Detection in Russian Texts*.
176. Wang, S., Yu, S. & Liu, P. Combining BERT and graph convolutional networks for enhanced fake review detection. in *Proceedings of 2025 International Conference on Artificial Intelligence and Computational Intelligence, AICI 2025* 24–28 (Association for Computing Machinery, Inc, 2025). doi:10.1145/3730436.3730440.
177. Castaneda, Á. H., Hernández, R. A. G., Ledeneva, Y. & Hernández, C. E. M. The impact of key ideas on automatic deception detection in text. *Computacion y Sistemas* **24**, 1229–1239 (2020).
178. Xu, T. & Rajivan, P. Determining psycholinguistic features of deception in phishing messages. *Information and Computer Security* **31**, 199–220 (2023).
179. Fornaciari, T., Poesio, M., Bianchi, F. & Hovy, D. *BERTective: Language Models and Contextual Information for Deception Detection*. (2021).
180. Azuma, D. *et al.* SVM, BERT, or LLM? A Comparative Study on Multilingual Instructed Deception Detection. *AI (Switzerland)* **6**, (2025).
181. Zeng, Z. Y. *et al.* A review structure based ensemble model for deceptive review spam. *Information (Switzerland)* **10**, (2019).
182. Saini, M. & Sharan, A. *Ensemble Learning to Find Deceptive Reviews Using Personality Traits and Reviews Specific Features E-Commerce, Online Shoping, E-Marketing, Opinion Mining*. *Journal of Digital Information Management* vol. 15 (2017).
183. Fusilier, D. H., Montes-Y-Gómez, M., Rosso, P. & Cabrera, R. G. Detection of opinion spam with character n-grams. in *Lecture Notes in Computer Science* vol. 9042 9042 LNCS 285–294 (Springer Verlag, 2015).
184. Pisarevskaya, D., Litvinova, T. & Litvinova, O. Deception Detection for the Russian Language: Lexical and Syntactic Parameters. in *International Conference Recent Advances in Natural Language Processing, RANLP* 1–10 (Incoma Ltd, 2017). doi:10.26615/978-954-452-038-0_001.
185. García-Galindo, M. del C., Hernández-Castañeda, Á., García-Hernández, R. A. & Ledeneva, Y. Identification of Deceptive Texts Using Cascade Classification. in *Lecture Notes in Computer Science*

*(including subseries Lecture Notes in Artificial Intelligence and Lecture Notes in Bioinformatics)* vol. 14755 LNCS 342–352 (Springer Science and Business Media Deutschland GmbH, 2024).
186. Ott, M., Choi, Y., Cardie, C. & Hancock, J. T. Finding Deceptive Opinion Spam by Any Stretch of the Imagination. in *Proceedings of the 49th annual meeting of the association for computational linguistics: Human language technologies* 309–319 (2011).
187. Hu, S. Detecting Concealed Information in Text and Speech. in *n Proceedings of the 57th annual meeting of the association for computational linguistics* 402–412 (2019).
188. Mohawesh, R., Bany Salameh, H., Jararweh, Y., Alkhalaileh, M. & Maqsood, S. Fake review detection using transformer-based enhanced LSTM and RoBERTa. *International Journal of Cognitive Computing in Engineering* **5**, 250–258 (2024).
189. de Ruiter, B. & Kachergis, G. The Mafiascum Dataset: A Large Text Corpus for Deception Detection. http://arxiv.org/abs/1811.07851 (2019).
190. Rubin, V. L. & Conroy, N. Discerning truth from deception: Human judgments and automation efforts. *First Monday* **17**, (2012).
191. Shahariar, G. M., Biswas, S., Omar, F., Shah, F. M. & Hassan, S. B. Spam Review Detection Using Deep Learning. in *2019 IEEE 10th Annual Information Technology, Electronics and Mobile Communication Conference (IEMCON)* 0027–0033 (IEEE, 2019).
192. Kao, Y. Y. *et al.* Detecting deceptive language in crime interrogation. in *Lecture Notes in Computer Science (including subseries Lecture Notes in Artificial Intelligence and Lecture Notes in Bioinformatics)* vol. 12204 LNCS 80–90 (Springer, 2020).
193. Glancy, F. H. & Yadav, S. B. A computational model for financial reporting fraud detection. *Decis. Support Syst.* **50**, 595–601 (2011).
194. Velutharambath, A., Sassenberg, K. & Klinger, R. What if Deception Cannot be Detected? A Cross-Linguistic Study on the Limits of Deception Detection from Text. http://arxiv.org/abs/2505.13147 (2026).
195. Abouelenien, M., Pérez-Rosas, V., Mihalcea, R. & Burzo, M. Deception detection using a multimodal approach. in *ICMI 2014 - Proceedings of the 2014 International Conference on Multimodal Interaction* 58–65 (Association for Computing Machinery, Inc, 2014). doi:10.1145/2663204.2663229.
196. Patra, C., Giri, D., Kundu, B., Maitra, T. & Wazid, M. Rhetorical Structure Theory-based machine intelligence-driven deceptive phishing attack detection scheme. *Journal of Information Security and Applications* **94**, (2025).
197. Hossain Sakib, M. K. *et al.* MVis4LD: Multimodal Visual Interactive System for Lie Detection. in *Lecture Notes in Computer Science (including subseries Lecture Notes in Artificial Intelligence and Lecture Notes in Bioinformatics)* vol. 14796 LNAI 28–43 (Springer Science and Business Media Deutschland GmbH, 2024).
198. Aadhavan, P., Vasantha Kumar, D., Suganthini Sri, R., Ganapathy, J. & Ramachandran, P. AI-Powered Phishing Detection in Email Forensics: A Machine Learning Approach for Cyber Threat. in *2025 10th International Conference on Frontiers of Signal Processing, ICFSP 2025* 146–150 (Institute of Electrical and Electronics Engineers Inc., 2025). doi:10.1109/ICFSP67350.2025.11353877.
199. Sánchez-Junquera, J., Villaseñor-Pineda, L., Montes-Y-Gómez, M. & Rosso, P. Character N-grams for detecting deceptive controversial opinions. in *Lecture Notes in Computer Science* vol. 11018 LNCS 135–140 (Springer Verlag, 2018).
200. Zhang, W., Jiang, Y. & Yoshida, T. Deep context identification of deceptive reviews using word vectors. in *Communications in Computer and Information Science* vol. 660 213–224 (Springer Verlag, 2016).
201. Wu, J. & Liu, Y. Deception detection methods incorporating discourse network metrics in synchronous computer-mediated communication. *J. Inf. Sci.* **46**, 64–81 (2020).
202. Zhang, H., Ding, Y., Cao, L., Wang, X. & Feng, L. Fine-Grained Question-Level Deception Detection via Graph-Based Learning and Cross-Modal Fusion. *IEEE Transactions on Information Forensics and Security* **17**, 2452–2467 (2022).
203. Almela, Á., Valencia-García, R. & Cantos, P. Seeing through deception: A computational approach to deceit detection in written communication. in *Proceedings of the workshop in computational approaches to deception detection* 15–22 (2012).
204. Phukon, P., Potikas, P. & Potika, K. Detecting Fake Reviews Using Aspect-Based Sentiment Analysis and Graph Convolutional Networks. *Applied Sciences (Switzerland)* **15**, (2025).
205. Fahfouh, A., Riffi, J., Mahraz, M. A., Yahyaouy, A. & Tairi, H. A Contextual Relationship Model for Deceptive Opinion Spam Detection. *IEEE Trans. Neural Netw. Learn. Syst.* **35**, 1228–1239 (2024).
206. Loconte, R., Russo, R., Capuozzo, P., Pietrini, P. & Sartori, G. Verbal lie detection using Large Language Models. *Sci. Rep.* **13**, (2023).
207. Mewada, A., Maurya, S. K. & Ansari, M. A. Comparative Study of Artificial Intelligence Approaches in Deceptive Opinion Detection. in *2025 2nd International Conference on Computational Intelligence,*

*Communication Technology and Networking, CICTN 2025* 999–1004 (Institute of Electrical and Electronics Engineers Inc., 2025). doi:10.1109/CICTN64563.2025.10932501.
208. Lin, C. H., Hsu, P. Y., Cheng, M. S., Lei, H. T. & Hsu, M. C. Identifying deceptive review comments with rumor and lie theories. in *Lecture Notes in Computer Science (including subseries Lecture Notes in Artificial Intelligence and Lecture Notes in Bioinformatics)* vol. 10386 LNCS 412–420 (Springer Verlag, 2017).
209. Maurya, S. K., Singh, D. & Maurya, A. K. Deceptive opinion spam detection using feature reduction techniques. *International Journal of System Assurance Engineering and Management* **15**, 1210–1230 (2024).
210. Dong, W. *et al. THE DETECTION OF FRAUDULENT FINANCIAL STATEMENTS: AN INTEGRATED LANGUAGE MODEL*. http://aisel.aisnet.org/pacis2014http://aisel.aisnet.org/pacis2014/383 (2014).
211. Rajoo, S. *et al.* Detecting false messages in the smartphone fault reporting system. in *Advances in Intelligent Systems and Computing* vol. 1073 759–768 (Springer, 2020).
212. Elmogy, A. M., Tariq, U., Ibrahim, A. & Mohammed, A. Fake Reviews Detection using Supervised Machine Learning. *International Journal of Advanced Computer Science and Applications* **12**, 2021 (2021).
213. Mendels, G., Levitan, S. I., Lee, K. Z. & Hirschberg, J. Hybrid acoustic-lexical deep learning approach for deception detection. in *Proceedings of the Annual Conference of the International Speech Communication Association, INTERSPEECH* vols 2017-August 1472–1476 (International Speech Communication Association, 2017).
214. Chou, H.-C., Chien, W.-S., Juan, D.-C. & Lee, C.-C. 'Does it Matter When I Think You Are Lying?' Improving Deception Detection by Integrating Interlocutor's Judgements in Conversations. in *Findings of the Association for Computational Linguistics: ACL-IJCNLP 2021* 1846–1860 (2021).
215. Saranya, G. *et al.* AI-Powered Phishing Detection: A Data-Driven Cybersecurity Approach. in *2025 International Conference on Data Science, Agents and Artificial Intelligence, ICDSAAI 2025* (Institute of Electrical and Electronics Engineers Inc., 2025). doi:10.1109/ICDSAAI65575.2025.11011572.
216. Ardulov, V., Durante, Z., Williams, S., Lyon, T. & Narayanan, S. Identifying truthful language in child interviews. in *ICASSP 2020-2020 IEEE International Conference on Acoustics, Speech and Signal Processing (ICASSP)* 8074–8078 (IEEE, 2020).
217. Fornaciari, T., Cagnina, L., Rosso, P. & Poesio, M. Fake opinion detection: how similar are crowdsourced datasets to real data? *Lang. Resour. Eval.* **54**, 1019–1058 (2020).
218. Aghakhani, H., MacHiry, A., Nilizadeh, S., Kruegel, C. & Vigna, G. Detecting deceptive reviews using generative adversarial networks. in *Proceedings - 2018 IEEE Symposium on Security and Privacy Workshops, SPW 2018* 89–95 (Institute of Electrical and Electronics Engineers Inc., 2018). doi:10.1109/SPW.2018.00022.
219. Salvetti, F., Lowe, J. B. & Martin, J. H. A Tangled Web: The Faint Signals of Deception in Text Boulder Lies and Truth Corpus (BLT-C). in *Proceedings of the Tenth International Conference on Language Resources and Evaluation (LREC'16)* 3510–3517 (2016).
220. Krishnan, A. Exploring Machine Learning and Transformer-based Approaches for Deceptive Text Classification: A Comparative Analysis. http://arxiv.org/abs/2308.05476 (2023).
221. Cocarascu, O. & Toni, F. Combining Deep Learning and Argumentative Reasoning for the Analysis of Social Media Textual Content Using Small Data Sets. https://doi.org/10.1162/coli (2018) doi:10.1162/coli.
222. Zhang, D., Zhou, L., Kehoe, J. L. & Kilic, I. Y. What Online Reviewer Behaviors Really Matter? Effects of Verbal and Nonverbal Behaviors on Detection of Fake Online Reviews. *Journal of Management Information Systems* **33**, 456–481 (2016).
223. Krishnan, A. B. H. Unmasking Falsehoods in Reviews: An Exploration of NLP Techniques. http://arxiv.org/abs/2307.10617 (2023).
224. Banerjee, S., Chua, A. Y. & Kim, J. J. Distinguishing between Authentic and Fictitious User-generated Hotel Reviews. in *2015 6th International Conference on Computing, Communication and Network Technologies* (2015).
225. Fuller, C. M., Biros, D. P. & Wilson, R. L. Decision support for determining veracity via linguistic-based cues. *Decis. Support Syst.* **46**, 695–703 (2009).
226. Wang, T. & Zhu, H. Voting for Deceptive Opinion Spam Detection. http://arxiv.org/abs/1409.4504 (2014).
227. Nguyen, T., Abri, F., Namin, A. S. & Jones, K. S. Deception and Lie Detection Using Reduced Linguistic Features, Deep Models and Large Language Models for Transcribed Data. in *Proceedings - 2024 IEEE 48th Annual Computers, Software, and Applications Conference, COMPSAC 2024* 376–381 (Institute of Electrical and Electronics Engineers Inc., 2024). doi:10.1109/COMPSAC61105.2024.00059.
228. Newman, M. L., Pennebaker, J. W., Berry, D. S. & Richards, J. M. Lying Words: Predicting Deception From Linguistic Styles. *Pers. Soc. Psychol. Bull.* **29**, 665–675 (2003).
229. Vogler, N. & Pearl, L. Using linguistically defined specific details to detect deception across domains. *Nat. Lang. Eng.* **26**, 349–373 (2020).

230. Heredia, B., Khoshgoftaar, T. M., Prusa, J. & Crawford, M. An Investigation of Ensemble Techniques for Detection of Spam Reviews. https://doi.org/10.1109/ICMLA.2016.38 (2016) doi:10.1109/ICMLA.2016.38.
231. Ludwig, S., van Laer, T., de Ruyter, K. & Friedman, M. Untangling a Web of Lies: Exploring Automated Detection of Deception in Computer-Mediated Communication. *Journal of Management Information Systems* **33**, 511–541 (2016).
232. Velutharambath, A., Wührl, A. & Klinger, R. How Entangled is Factuality and Deception in German? in *Findings of the Association for Computational Linguistics: EMNLP 2024* 9538–9554 (2024).
233. Mbaziira, A. V. & Murphy, D. R. An empirical study on detecting deception and cybercrime using artificial neural networks. in *ACM International Conference Proceeding Series* 42–46 (Association for Computing Machinery, 2018). doi:10.1145/3193077.3193080.
234. Hernández-Castañeda, Á. & Calvo, H. Deceptive text detection using continuous semantic space models. in *Intelligent Data Analysis* vol. 21 679–695 (IOS Press, 2017).
235. Alzghoul, J. R., Alzghool, M. & Abdallah, E. E. A new approach for deception detection in open domain text. *International Journal of Business Analytics* **8**, 1–13 (2021).
236. Humpherys, S. L., Moffitt, K. C., Burns, M. B., Burgoon, J. K. & Felix, W. F. Identification of fraudulent financial statements using linguistic credibility analysis. *Decis. Support Syst.* **50**, 585–594 (2011).
237. Fornaciari, T., Celli, F. & Poesio, M. The effect of personality type on deceptive communication style. in *Proceedings - 2013 European Intelligence and Security Informatics Conference, EISIC 2013* 1–6 (2013). doi:10.1109/EISIC.2013.8.
238. Mewada, A. *et al.* Deceptive Opinion Detection Using Stacking-Based Deep Ensemble Learning. in *ICDT 2025 - 3rd International Conference on Disruptive Technologies* 1614–1617 (Institute of Electrical and Electronics Engineers Inc., 2025). doi:10.1109/ICDT63985.2025.10986298.
239. Cagnina, L. C. & Rosso, P. *Classification of Deceptive Opinions Using a Low Dimensionality Representation*. http://text-processing.com/demo/sentiment/ (2015).
240. Narayan, R., Rout, J. K. & Jena, S. K. Review spam detection using opinion mining. in *Advances in Intelligent Systems and Computing* vol. 519 273–279 (Springer Verlag, 2018).
241. Shojaee, S., Murad, M. A. A., Azman, A. Bin, Sharef, N. M. & Nadali, S. Detecting deceptive reviews using lexical and syntactic features. in *International Conference on Intelligent Systems Design and Applications, ISDA* 53–58 (IEEE Computer Society, 2014). doi:10.1109/ISDA.2013.6920707.
242. Vy Nguyen, N. T., Childress, F. D. & Yin, Y. Debate-Driven Multi-Agent LLMs for Phishing Email Detection. in *ISDFS 2025 - 13th International Symposium on Digital Forensics and Security* (Institute of Electrical and Electronics Engineers Inc., 2025). doi:10.1109/ISDFS65363.2025.11012014.
243. Verhoeven, B. & Daelemans, W. CLiPS Stylometry Investigation (CSI) corpus: A Dutch corpus for the detection of age, gender, personality, sentiment and deception in text. in *LREC 2014-Ninth Intenational Conference on Language Resources and Evaluation* 3081–3085 (2014).
244. Velutharambath, A. & Klinger, R. UNIDECOR: A Unified Deception Corpus for Cross-Corpus Deception Detection. in *Proceedings of the Annual Meeting of the Association for Computational Linguistics* 39–51 (Association for Computational Linguistics (ACL), 2023). doi:10.18653/v1/2023.wassa-1.5.
245. Tanwar, P. & Rai, P. A proposed system for opinion mining using machine learning, nlp and classifiers. *IAES International Journal of Artificial Intelligence* **9**, 726–733 (2020).
246. Markowitz, D. M. & Hancock, J. T. Linguistic Traces of a Scientific Fraud: The Case of Diederik Stapel. *PLoS One* **9**, e105937 (2014).
247. Anass, F., Jamal, R., Mahraz, M. A., Ali, Y. & Tairi, H. Deceptive Opinion Spam based on Deep Learning. in *2020 Fourth International Conference On Intelligent Computing in Data Sciences (ICDS)* 1–5 (IEEE, 2020).
248. Du, X., Zhao, F., Zhu, Z. & Han, P. DRDF: A Deceptive Review Detection Framework of Combining Word-Level, Chunk-Level, and Sentence-Level Topic-Sentiment Models. in *Proceedings of the International Joint Conference on Neural Networks* vols 2021-July (Institute of Electrical and Electronics Engineers Inc., 2021).
249. Mihalcea, R. & Strapparava, C. *The Lie Detector: Explorations in the Automatic Recognition of Deceptive Language*. *ACL and AFNLP* (2009).
250. Capuozzo, P., Lauriola, I., Strapparava, C., Aiolli, F. & Sartori, G. *DecOp: A Multilingual and Multi-Domain Corpus For Detecting Deception In Typed Text*. (2020).
251. Mihalcea, R., Pérez-Rosas, V. & Burzo, M. Automatic detection of deceit in verbal communication. in *ICMI 2013 - Proceedings of the 2013 ACM International Conference on Multimodal Interaction* 131–134 (2013). doi:10.1145/2522848.2522888.
252. Minhas, S. & Hussain, A. From Spin to Swindle: Identifying Falsification in Financial Text. *Cognit. Comput.* **8**, 729–745 (2016).
253. Veluru, S. R., Teja Erukude, S. & Marella, V. C. Multimodal Detection of Fake Reviews using BERT and ResNet-50. in *Proceedings of the 4th International Conference on Innovative Mechanisms for Industry*

*Applications, ICIMIA 2025* 877–882 (Institute of Electrical and Electronics Engineers Inc., 2025). doi:10.1109/ICIMIA67127.2025.11200892.
254. Patel, R. & Thakkar, P. Opinion Spam Detection Using Feature Selection. https://doi.org/10.1109/.127 (2014) doi:10.1109/.127.
255. Guarasci, R., Catelli, R. & Esposito, M. Classifying deceptive reviews for the cultural heritage domain: A lexicon-based approach for the Italian language. *Expert Syst. Appl.* **252**, (2024).
256. Akkol, E. & Gökşen, Y. Deception Detection in Turkish Hotel Reviews: A Comparative Study of Machine Learning and Deep Learning Approaches. *ACTA INFOLOGICA* https://doi.org/10.26650/acin.1606539 (2025) doi:10.26650/acin.1606539.
257. Burgoon, J. K. Predicting Veracity From Linguistic Indicators. *J. Lang. Soc. Psychol.* **37**, 603–631 (2018).
258. Fornaciari, T. & Poesio, M. On the Use of Homogenous Sets of Subjects in Deceptive Language Analysis. in *Proceedings of the EACL 2012 Workshop on Computational Approaches to Deception Detection* 39–47 (Association for Computational Linguistics, 2012).
259. Masip, J., Bethencourt, M., Lucas, G., Segundo, M. S. S. & Herrero, C. Deception detection from written accounts. *Scand. J. Psychol.* **53**, 103–111 (2012).
260. Bond, G. D. & Lee, A. Y. Language of lies in prison: Linguistic classification of prisoners' truthful and deceptive natural language. *Appl. Cogn. Psychol.* **19**, 313–329 (2005).
261. Morales, M. R., Scherer, S. & Levitan, R. OpenMM: An open-source multimodal feature extraction tool. in *Proceedings of the Annual Conference of the International Speech Communication Association, INTERSPEECH* vols 2017-August 3354–3358 (International Speech Communication Association, 2017).
262. Zhang, W., Wang, Q., Li, X., Yoshida, T. & Li, J. DCWord: A Novel Deep Learning Approach to Deceptive Review Identification by Word Vectors. *J. Syst. Sci. Syst. Eng.* **28**, 731–746 (2019).
263. Salloum, S., Tahat, K., Mansoori, A., Alfaisal, R. & Tahat, D. Analysis of Fraudulent Job Postings Using Machine Learning. in *2024 International Conference on Intelligent Computing, Communication, Networking and Services, ICCNS 2024* 268–270 (Institute of Electrical and Electronics Engineers Inc., 2024). doi:10.1109/ICCNS62192.2024.10776527.
264. Akbar, A. H., Siagian, M. & Aritsugi, M. Exploiting Function Words Feature in Classifying Deceptive and Truthful Reviews. in *Aritsugi, M. (2018, September). Exploiting function words feature in classifying deceptive and truthful reviews. In 2018 Thirteenth International Conference on Digital Information Management (ICDIM)* 51–56 (IEEE, 2018).
265. Matsumoto, D. & Hwang, H. C. Differences in Word Usage by Truth Tellers and Liars in Written Statements and an Investigative Interview After a Mock Crime. *Journal of Investigative Psychology and Offender Profiling* **12**, 199–216 (2015).
266. Evans, A. D. *et al.* The Effects of Repetition on Children's True and False Reports. *Psychiatry, Psychology and Law* **19**, 517–529 (2012).
267. Pisarevskaya, D., Galitsky, B., Ozerov, A. & Taylor, J. An anatomy of a lie: Discourse patterns in customer complaints deception dataset. in *The Web Conference 2019 - Companion of the World Wide Web Conference, WWW 2019* 373–380 (Association for Computing Machinery, Inc, 2019). doi:10.1145/3308560.3316468.
268. Fuller, C. M., Biros, D. P. & Delen, D. Exploration of feature selection and advanced classification models for high-stakes deception detection. in *Proceedings of the Annual Hawaii International Conference on System Sciences* (2008). doi:10.1109/HICSS.2008.158.
269. Bajaj, N. *et al.* Fraud detection in telephone conversations for financial services using linguistic features. http://arxiv.org/abs/1912.04748 (2019).
270. Oh, Y. W. & Park, C. H. Machine Cleaning of Online Opinion Spam: Developing a Machine-Learning Algorithm for Detecting Deceptive Comments. *American Behavioral Scientist* **65**, 389–403 (2021).
271. Panda, S. & Levitan, S. I. Improving Cross-domain, Cross-lingual and Multi-modal Deception Detection. in *Proceedings of the 60th Annual Meeting of the Association for Computational Linguistics: Student Research Workshop* 383–390 (2022).
272. Li, J., Ott, M., Cardie, C. & Hovy, E. Towards a General Rule for Identifying Deceptive Opinion Spam. in *Proceedings of the 52nd Annual Meeting of the Association for Computational Linguistics* 1566–1576 (Association for Computational Linguistics, 2014).
273. Dykstra, V. W., Lyon, T. D. & Evans, A. D. Maltreated and non-maltreated children's truthful and dishonest reports: Linguistic and syntactic differences. *Front. Psychol.* **13**, (2022).
274. Burgoon, J. K., Blair, J. P., Qin, T. & Nunamaker, J. F. *Detecting Deception through Linguistic Analysis*. *LNCS* vol. 2665 (2003).
275. Hosomi, N., Sakti, S., Yoshino, K. & Nakamura, S. Deception Detection and Analysis in Spoken Dialogues based on FastText. in *2018 Asia-Pacific Signal and Information Processing Association Annual Summit and Conference (APSIPA ASC)* 139–142 (IEEE, 2018).

276. Ziani, A. *et al.* Deceptive Opinions Detection Using New Proposed Arabic Semantic Features. in *Procedia CIRP* vol. 189 29–36 (Elsevier B.V., 2021).
277. Krishnamurthy, G., Majumder, N., Poria, S. & Cambria, E. A deep learning approach for multimodal deception detection. in *International Conference on Computational Linguistics and Intelligent Text Processing* (ed. Gelbukh, A.) vol. 13396 87–96 (Springer Nature Switzerland, Cham, 2018).
278. Kim, S., Chang, H., Lee, S., Yu, M. & Kang, J. Deep semantic frame-based deceptive opinion spam analysis. in *International Conference on Information and Knowledge Management, Proceedings* vols 19-23-Oct-2015 1131–1140 (Association for Computing Machinery, 2015).
279. Twitchell, D. P. *et al. Automated Determination of the Veracity of Interview Statements from People of Interest to an Operational Security Force*. (2006).
280. Wang, J. H. & Liu, Y. H. Feature extraction for product advertising reviews identification in social media. in *ACM International Conference Proceeding Series* vols 07-09-Ocobert-2015 (Association for Computing Machinery, 2015).
281. Birim, Ş. Ö. *et al.* Detecting fake reviews through topic modelling. *J. Bus. Res.* **149**, 884–900 (2022).
282. Papantoniou, K. *et al.* Deception detection in text and its relation to the cultural dimension of individualism/collectivism. *Nat. Lang. Eng.* **28**, 545–606 (2022).
283. Bhatia, T. & Meena, J. Detection of Fake Online Recruitment Using Machine Learning Techniques. in *Proceedings - 2022 4th International Conference on Advances in Computing, Communication Control and Networking, ICAC3N 2022* 300–304 (Institute of Electrical and Electronics Engineers Inc., 2022). doi:10.1109/ICAC3N56670.2022.10074276.
284. Forsyth, L. & Anglim, J. Using text analysis software to detect deception in written short-answer questions in employee selection. *International Journal of Selection and Assessment* **28**, 236–246 (2020).
285. Mendes, P., Maia, E. & Praça, I. MeAJOR Corpus: A Multi-Source Dataset for Phishing Email Detection. http://arxiv.org/abs/2507.17978 (2025).
286. Salunkhe, A. Attention-based Bidirectional LSTM for Deceptive Opinion Spam Classification. http://arxiv.org/abs/2112.14789 (2021).
287. Levitan, S. I. *et al.* Combining Acoustic-Prosodic, Lexical, and Phonotactic Features for Automatic Deception Detection. in *Proc. Interspeech 2016* 2006–2010 (2016).
288. Şenol, A., Agrawal, G. & Liu, H. Domain Knowledge-Enhanced LLMs for Fraud and Concept Drift Detection. *Electronics (Switzerland)* **15**, (2026).
289. Monisha, G. B. & Nayak, J. S. Detection of Online Fake Review Using Deep Learning. in *Lecture Notes in Networks and Systems* vol. 948 LNNS 161–172 (Springer Science and Business Media Deutschland GmbH, 2024).

# Supplementary Information

## Supplementary Results

Study identification and selection
The adapted version of the PRISMA flow diagram (*87*) summarizes the reports identification and selection process (Figure S1). Starting from a search of 29,023 reports, screened with the help of an active-learning screening tool (ASReview LAB v.1, https://asreview.nl), a total of 289 reports qualified for inclusion, yielding to the coding of 6,136 models. Of the 289 reports, 47.06% (*n*=136) were peer-reviewed journal articles, 46.71% (*n*=135) consisted of conference papers, and 6.23% (*n*=18) were preprints. The number of included reports per publication year spanned from 1982 to 2025, with an increased number of publications in the last 15 years (Figure S2).

Type of deception, data origin, and research design interplay
When examining the interplay between the type of deception, data origin, and research design, Table S1 reveals that offline experimental designs addressed the widest range of deception types, including concealment, embedded lies, falsification, and social desirability, although fabrication remained the most frequently investigated. In contrast, quasi-experimental designs focused almost exclusively on fabrication, while naturalistic designs predominantly examined fabrication and mixed forms of deception. Overall, experimental approaches appeared to be the most versatile research design for covering different types of deception. Details on the frequencies and percentages of deception types, topics of investigation, ground-truth specification, and data sources are reported in Table S2.

Dataset size
Dataset sizes varied substantially across sources and research designs (Tables S3 and S4). Overall, offline sources yielded smaller datasets compared to online sources, reflecting how online platforms enable large-scale studies (see Douglas et al., 2023). Among online sources, naturalistic designs that scraped data from platforms (e.g., Facebook, Amazon, Yelp) produced larger datasets (Median=8,897.50; range: 103–2,542,553). For offline sources, the largest dataset was from one report from a quasi-experimental design (Median=856), followed by datasets from experimental studies (Median=277; range: 20–9,104), albeit much smaller than those collected from online sources. This wide range highlights the heterogeneity of resources used in deception detection research, spanning small-scale controlled experiments to large-scale online data, which likely reflect differences in research goals, feasibility, and data accessibility.

Computational choices at the model level
At the model level, the number of models using each type of textual representation, model category, evaluation procedure, and performance metric is reported in Table S5. This breakdown shows how frequently each methodological choice was adopted across the included models, complementing the dataset-level statistics reported above.

Pooled accuracy estimates
After applying the eligibility criteria for the quantitative synthesis, the final meta-analysis included 3,653 classification models nested within 97 datasets and 161 independent reports. The pooled effect was estimated on the logit scale, controlling for dataset class imbalance by including the proportion of the majority class as a continuous moderator centered at 0.50. This effect was statistically significant ($\beta = 1.07$, $SE = 0.08$, $z = 12.74$, $p < .001$, 95% CI [0.90, 1.23]) and corresponded to a predicted pooled accuracy of 74.43% ([71.17%-77.43%]; see Figure 2). Substantial residual heterogeneity remained after accounting for class imbalance, $QE(3{,}651) = 468{,}605.51$, $p < .001$, $I^2 = 99.6\%$, indicating that approximately 99.6% of the total variability in the performance estimates was attributable to residual differences in the underlying effects rather than sampling error. Of this total variability, 34.8% was attributable to residual heterogeneity between datasets ($\sigma^2 = 0.29$, $SD = 0.54$, [0.19,0.46], $I^2 = 34.8$) and 64.8% between reports ($\sigma^2 = 0.55$, $SD = 0.74$, [0.42, -][10], $I^2 = 64.8$). Further, the multilevel random-effects meta-regression showed that dataset class imbalance significantly influenced reported classification accuracy, $QM(1) = 4187.96$, $p < .001$, with a higher proportion of the majority class associated with higher reported accuracy ($\beta = 1.40$, SE = 0.02, z = 64.71, p < .001, [1.36, 1.44]). Back-transformed predicted accuracies rose from 77.0% at a majority-class proportion of 0.60 to 83.6% at 0.90. Notably, this gain trails what a trivial classifier would achieve by always predicting the majority class (60% and 90% accuracy, respectively): rather than inflating performance, models appear to underexploit class imbalance, increasingly so as imbalance grows.

[10] The upper profile-likelihood confidence bound was not estimable due to convergence issues.

Analysis of moderators

A second multilevel meta-regression examined the influence of methodological characteristics while controlling for dataset class imbalance. Dataset-level residual heterogeneity accounted for the 87.2% of the total variability ($\sigma^2 = 2.73$, *SD* = 1.65, [1.99, 3.80], $I^2 = 87.2$), suggesting that dataset-specific characteristics not fully captured by the coded moderators account for a large portion of the residual variance in reported accuracy. Report-level residual heterogeneity accounted for a further 12.6% of variability ($\sigma^2 = 0.40$, *SD* = 0.63, [0.30, 0.53], $I^2 = 12.6$), indicating that variability in modeling choices within the same dataset contributes comparatively less to differences in reported performance. Together, these two levels of heterogeneity accounted for 99.8% of the total variability in reported accuracy beyond sampling error.

Regarding the fixed effects of methodological choices, the model explained 30.1% of the total variance in reported accuracy (marginal $R^2 = 0.30$), $QM(20) = 22038.02$, $p < .001$, although considerable residual heterogeneity remained ($QE(3632) = 325084.72$, $p < .001$). When both fixed and random effects are considered jointly, the full model accounted for 91.2% of the total variance (conditional $R^2 = 0.91$). Full details of the moderator analysis are reported in Table 2 in the main text.

First, the data source significantly moderated accuracy ($QM(6) = 210.29$, $p < .001$). Compared to offline experiments, offline quasi-experimental data showed a modest but significantly higher accuracy over offline experiments ($\beta = 0.92$, $SE = 0.46$, $z = 2.00$, $p = .046$), while offline naturalistic data did not differ significantly from the reference category ($\beta = 0.78$, $SE = 0.59$, $z = 1.33$, $p = .183$). In contrast, online quasi-experimental data showed larger advantage ($\beta = 1.59$, $SE = 0.46$, $z = 3.45$, $p < .001$), followed by online naturalistic data ($\beta = 1.42$, $SE = 0.46$, $z = 3.07$, $p = .002$), and online experiments ($\beta = 1.17$, $SE = 0.46$, $z = 2.55$, $p = .011$). Notably, mixed data sources showed the highest accuracy ($\beta = 3.88$, $SE = 0.78$, $z = 4.99$, $p < .001$). Taken together, these results suggest that mixed and online data sources are associated with systematically higher accuracy estimates than offline sources, which may reflect differences in sample size and ecological validity.

Ground-truth operationalization also significantly affected accuracy ($QM(3) = 2922.77$, $p < .001$). Compared with a clear and verifiable ground truth, reports relying on a directly inferred ($\beta = -0.35$, $SE = 0.10$, $z = -3.47$, $p = .0005$) or indirectly inferred ground-truth ($\beta = -3.71$, $SE = 0.11$, $z = -34.36$, $p < .001$) showed significantly lower reported accuracies, whereas reports using clear but non-verifiable ground truth reported higher accuracies ($\beta = 0.48$, $SE = 0.08$, $z = 6.33$, $p < .001$). These findings indicate that lower-quality ground truth tends to inflate accuracy, whereas inferred ground truth levels significantly reduce models' accuracy.

Text-representation was a strong predictor of model performance ($QM(4) = 5520.63$, $p < .001$). Relative to linguistic and statistical features, embedding-based ($\beta = 0.12$, $SE = 0.01$, $z = 19.68$, $p < .001$), term-frequency ($\beta = 0.25$, $SE = 0.005$, $z = 54.62$, $p < .001$), and topic-based representations ($\beta = 0.11$, $SE = 0.01$, $z = 8.75$, $p < .001$) were all associated with significantly higher classification accuracy, with hybrid representations ($\beta = 0.33$, $SE = 0.005$, $z = 67.46$, $p < .001$) yielding the largest positive effect. Model architecture also significantly predicted accuracy ($QM(4) = 1356.37$, $p < .001$). Transformer-based models showed the highest accuracy relative to traditional statistical learning approaches ($\beta = 0.32$, $SE = 0.01$, $z = 31.41$, $p < .001$), followed by ensemble models ($\beta = 0.05$, $SE = 0.005$, $z = 8.96$, $p < .001$), and neural networks ($\beta = 0.03$, $SE = 0.005$, $z = 5.67$, $p < .001$). Notably, large language models showed slightly lower reported accuracies ($\beta = -0.09$, $SE = 0.02$, $z = -5.49$, $p < .001$). Finally, the evaluation procedure significantly influenced reported performance ($QM(2) = 1866.72$, $p < .001$). Models evaluated on independent datasets reported substantially lower accuracies than those evaluated using within cross-validation ($\beta = -0.38$, $SE = 0.01$, $z = -39.04$, $p < .001$), whereas hold-out test sets slightly inflated accuracies than cross-validation after adjustment for the remaining moderators ($\beta = 0.06$, $SE = 0.01$, $z = 4.25$, $p < .001$).

Limitations

This systematic review and meta-analysis are subject to a few limitations that should be noted. First, the choices regarding eligibility criteria and search strategy inevitably shape the scope and comprehensiveness of our findings. For example, we excluded all research on fake reviews and misinformation because, although they constitute a form of deception (*92*), they represent a separate research line that would have unduly influenced our review. Second, some full texts, especially those found in conference proceedings, showed substandard reporting quality and were shorter in length. This may have introduced errors in the coding phase, particularly for variables that were rarely explicitly detailed, such as the type of deception or the ground truth, and that we often had to infer.

Furthermore, in using ASReview Lab to assist human re-viewers in screening an extensive body of literature, we may have inevitably introduced screening errors, disagreements, and excluded edge cases. However, we argue that the hunt for the very last paper remains a myth, as previous research has shown that, regardless of the methods employed (whether fully manual or assisted), some relevant papers will inevitably be missed (*93*). Finally, substantial heterogeneity was observed across reports and datasets, suggesting that the pooled performance was highly report-dependent and dataset-specific despite estimates coming from a large sample of models ($n$ = 3,653). Some combinations of levels were sparse or absent (e.g., indirectly inferred ground truth was concentrated almost entirely within online naturalistic data and was absent from offline research designs) and multilevel analysis could not fully separate the effects of such overlapping factors from dataset-level differences. Therefore, we recommend caution when interpreting our meta-analytical findings.

**Fig. S1.** Adapted version of the PRISMA flow diagram.

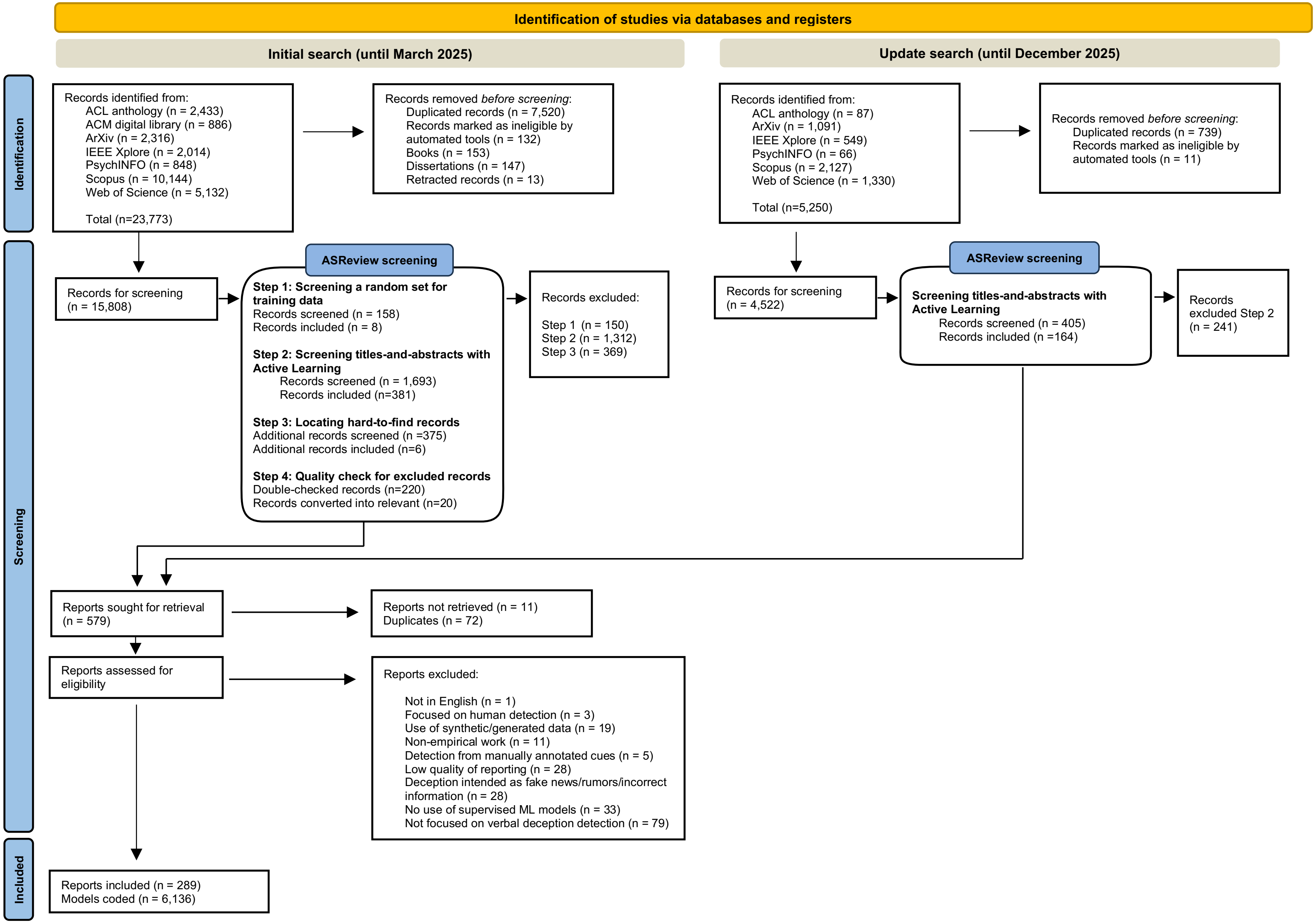

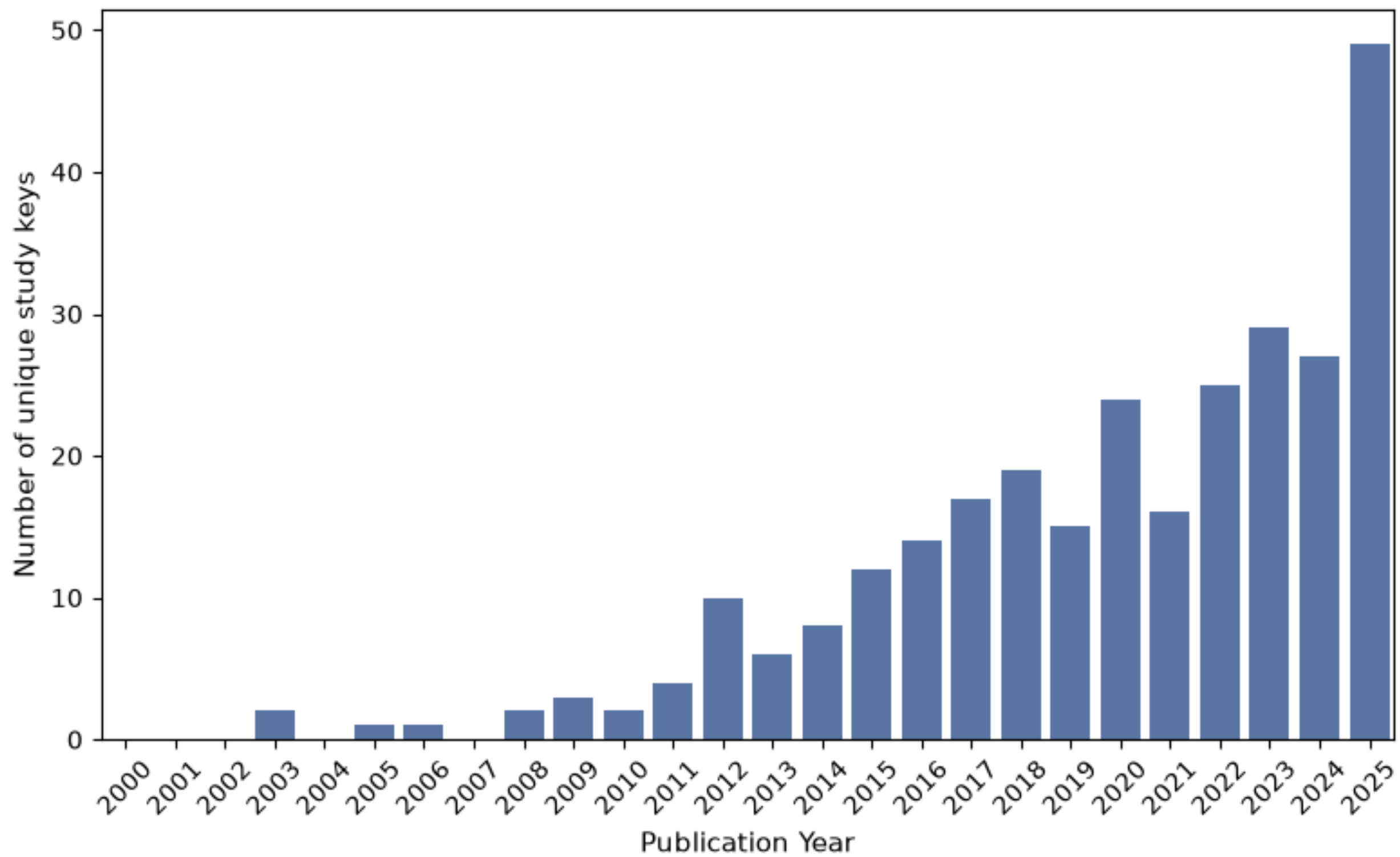

**Fig. S2** Distribution of published reports over time (2000-2025).
*Note.* One model from 1982, included in the systematic review, was excluded from the plot for a better visualization.

**Table S1.** Number of reports investigating different forms of deception by source and research design.

| * | Experimental | | Quasi-experimental | | Naturalistic | |
|---|---|---|---|---|---|---|
| | Offline | Online | Offline | Online | Offline | Online |
| Concealment | 2 | - | - | - | - | - |
| Embedded lies | 2 | 2 | - | - | 1 | - |
| Fabrication | 17 | 19 | 1 | 61 | 12 | 51 |
| Falsification | 2 | - | - | - | - | 1 |
| Social Desirability | 1 | - | - | - | - | - |
| Mixed | 18 | 5 | - | 1 | 39 | 13 |

*Reports ($n$=39) that relied on multiple sources of data collection were excluded from this table as the type of deception could not be derived. Similarly, reports ($n$=2) for which information on the research design for data collection was missing were excluded.

**Table S2.** Frequency and percentage of each level of the report-level coded variables among the included reports.

| Item category | Item label | Levels* | Frequency (%)† |
|---|---|---|---|
| **Deception operationalization** | Type of deception | Fabrication | 184 (63.67%) |
| | | Mixed | 94 (32.53%) |
| | | Embedded lies | 5 (1.73%) |
| | | Falsification | 3 (1.04%) |
| | | Concealment or omission | 2 (0.69%) |
| | | Social desirability | 1 (0.35%) |
| | Topic of investigation | Fake reviews | 120 (41.52%) |
| | | Past events | 46 (15.92%) |
| | | Mixed | 34 (11.76%) |
| | | Email/sms spam | 24 (8.30%) |
| | | Fake personal opinions | 19 (6.57%) |
| | | Identity and personal information | 19 (6.57%) |
| | | Scientific fraud | 8 (2.77%) |
| | | Cheating on a test | 5 (1.73%) |
| | | Open-domain | 5 (1.73%) |
| | | Personal feelings | 3 (1.04%) |
| | | Object descriptions | 3 (1.04%) |
| | | Future intentions and promises | 3 (1.04%) |
| | Ground truth | Clear and verifiable | 36 (12.46%) |
| | | Clear but not verifiable | 110 (38.06%) |
| | | Directly inferred | 38 (13.15%) |
| | | Indirectly inferred | 37 (12.80%) |
| | | Mixed | 43 (14.88%) |
| | | No information | 25 (8.65%) |
| **Dataset** | Source | Offline experiment | 42 (14.53%) |
| | | Offline quasi-experiment | 1 (0.35%) |
| | | Offline naturalistic data | 52 (17.99%) |
| | | Online experiment | 26 (9.00%) |
| | | Online quasi-experiment | 62 (21.45%) |
| | | Online naturalistic data | 65 (22.49%) |
| | | Mixed | 39 (13.45%) |
| | | No information | 2 (0.69%) |

*Type of deception and Topic of investigation were coded by first following a bottom-up, inductive approach and then recoded into overarching categories. Coding of Ground truth and Source followed a top-down, deductive approach, in which full texts were coded using pre-defined codes.
†Frequency values and percentages were computed at the report level (*n*=289)

**Table S3.** Descriptive statistics of no. of statements (*M*, *SD*) by source of data collection and research design.

| **Source** | **Design** | ***N* (%)*** | ***M*** | ***SD*** | ***Median*** | **Range** |
|---|---|---|---|---|---|---|
| | Experiment | 38 (13.15%) | 1,328.19 | 2122.58 | 277 | 20 - 9,104 |
| Offline | Quasi-experiment | 1 (0.35%) | 856 | - | 856 | 856 - 856 |
| | Naturalistic | 52 (17.99%) | 2,013.22 | 4,586.37 | 179 | 18 – 17,880 |
| | Experiment | 25 (8.65%) | 1,895.87 | 2,112.59 | 1,031.00 | 30.75 - 7,168 |
| Online | Quasi-experiment | 61 (21.11%) | 2,555.64 | 4,825.23 | 1,600 | 640 – 31,146.68 |
| | Naturalistic | 61 (21.11%) | 76,947.62 | 336,053.47 | 8,897.50 | 103 – 2,542,553 |
| Mixed | Mixed | 39 (13.49%) | 49,087.01 | 260,902.34 | 1414.24 | 121 – 1,630,263.25 |

* No. of statements was averaged per report before computing the descriptive statistics by source and design. Two reports were excluded because of unclear research designs and origin; ten reports were excluded due to missing information. *N* represents the number of reports that reported the number of statements in full text. Percentages refer to the proportion of all reports included in the systematic review.

**Table S4.** Descriptive statistics of no. of statements (M, SD) by source of data collection and research design at the model-level.

| **Origin** | **Design** | ***N* (%)*** | ***M*** | ***SD*** | ***Median*** | **Range** |
|---|---|---|---|---|---|---|
| Offline | Experiment | 843 (13.76%) | 546.58 | 1,147.87 | 226 | 20 - 9,104 |
| | Quasi-experiment | 4 (0.07%) | 1,328 | 545.02 | 1,328 | 856 - 1,800 |
| | Naturalistic data | 548 (8.95%) | 1,463.61 | 3,892.05 | 150 | 18 - 17,880 |
| Online | Experiment | 1,536 (25.07%) | 1096.59 | 1,698.35 | 500 | 14 - 14,343 |
| | Quasi-experiment | 2,067 (33.74%) | 3,591.01 | 27,575.97 | 1,600 | 200 - 501,472 |
| | Naturalistic data | 955 (15.59%) | 74,796.77 | 401,288.41 | 6,014 | 78 - 3,868,306 |
| Mixed | Mixed | 173 (2.82%) | 14,294.28 | 39,033.85 | 320 | 200 – 135,894 |

* Ten models with unclear data sources are not reported in this table. *N* represents the number of models from the included reports.

**Table S5.** Frequency and percentage of each level of the model-level coded variables among the included reports.

| Item category | Item label | Levels* | Frequency (%)† |
|---|---|---|---|
| **Linguistic features** | Text-representation | Linguistic and statistical features | 1,256 (20.47%) |
| | | Term frequencies | 1,737 (28.31%) |
| | | Embeddings | 1,893 (30.85%) |
| | | Hybrid approaches | 1,159 (18.89%) |
| | | Topic and latent semantic features | 69 (1.12%) |
| | | Unclear information | 22 (0.36%) |
| **Computational approach** | Model category | Statistical models | 2,973 (48.45%) |
| | | Ensemble models | 815 (13.28 %) |
| | | Neural networks | 1,119 (18.24%) |
| | | Transformer-based models | 600 (9.78%) |
| | | Large language models | 628 (10.23%) |
| | | Unclear information | 1 (0.02%) |
| | Evaluation | In-sample evaluation | 29 (0.47%) |
| | | Hold-out test-set | 1590 (25.91%) |
| | | Within cross-validation | 2871 (46.79%) |
| | | Different dataset | 1470 (23.96%) |
| | | Unclear information | 176 (2.87%) |
| | Metrics | Accuracy | 2572 (41.92%) |
| | | F1 score | 1257 (20.49%) |
| | | Both (accuracy and F1) | 2073 (33.78%) |
| | | Other metrics | 234 (3.81%) |

*Each level was coded by first following a bottom-up, inductive approach and then recoded into overarching categories.
†Frequency values and percentages were computed at the model level ($n$=6,136)

## Data S1. (separate file)

Reference list of key reports from previous meta-analysis (Hauch et al., 2015) and systematic reviews (Costancio et al., 2022) to assess the search and title-and-abstract screening quality.

## Data S2. (separate file)

Reference list and details of reused datasets that were reported in an academic outlet and fit within the first three levels of ground truth.